\documentclass[sigconf]{acmart}
\usepackage{algorithm}
\usepackage{algpseudocode}
\usepackage{fancyvrb}
\usepackage{multirow}
\usepackage{multicol}
\AtBeginDocument{%
  \providecommand\BibTeX{{%
    \normalfont B\kern-0.5em{\scshape i\kern-0.25em b}\kern-0.8em\TeX}}}

\setcopyright{acmcopyright}
\copyrightyear{2022}
\acmYear{2022}
\setcopyright{rightsretained}
\acmConference[GECCO '22]{Genetic and Evolutionary Computation Conference}{July 9--13, 2022}{Boston, MA, USA}
\acmBooktitle{Genetic and Evolutionary Computation Conference (GECCO '22), July 9--13, 2022, Boston, MA, USA}
\acmDOI{10.1145/3512290.3528725}
\acmISBN{978-1-4503-9237-2/22/07}

\newcommand{\red}[1]{\textcolor{red}{#1}}

\begin{document}
\title{High-performance Evolutionary Algorithms for Online Neuron Control}


\author{Binxu Wang}
\affiliation{%
  \institution{Harvard Medical School}
  \streetaddress{220 Longwood Ave}
  \city{Boston}
  \state{MA}
  \country{USA}
  \postcode{02115}}
\additionalaffiliation{%
  \institution{Washington University in St Louis}
  \streetaddress{4559 Scott Ave}
  \city{St Louis}
  \state{MO}
  \country{USA}
  \postcode{63110}}
\email{binxu_wang@hms.harvard.edu}

\author{Carlos R. Ponce}
\affiliation{%
  \institution{Harvard Medical School}
  \streetaddress{220 Longwood Ave}
  \city{Boston}
  \state{MA}
  \country{USA}
  \postcode{02115}}
\email{carlos@hms.harvard.edu}

\renewcommand{\shortauthors}{Wang and Ponce}

\begin{abstract}
Recently, optimization has become an emerging tool for neuroscientists to study neural code. In the visual system, neurons respond to images with graded and noisy responses. Image patterns eliciting highest responses are diagnostic of the coding content of the neuron. To find these patterns, we have used black-box optimizers to search a 4096d image space, leading to the evolution of images that maximize neuronal responses. Although genetic algorithm (GA) has been commonly used, there haven't been any systematic investigations to reveal the best performing optimizer or the underlying principles necessary to improve them.

Here\footnotemark, we conducted a large scale \textit{in silico} benchmark of optimizers for activation maximization and found that Covariance Matrix Adaptation (CMA) excelled in its achieved activation. We compared CMA against GA and found that CMA surpassed the maximal activation of GA by 66\% \textit{in silico} and 44\% \textit{in vivo}. We analyzed the structure of Evolution trajectories and found that the key to success was not covariance matrix adaptation, but local search towards informative dimensions and an effective step size decay. Guided by these principles and the geometry of the image manifold, we developed SphereCMA optimizer which competed well against CMA, proving the validity of the identified principles.
\footnotetext{Code available at \url{https://github.com/Animadversio/ActMax-Optimizer-Dev}}

\end{abstract}



\begin{teaserfigure}
\vspace{-10pt}
  \centering
  \includegraphics[width=.8\textwidth]{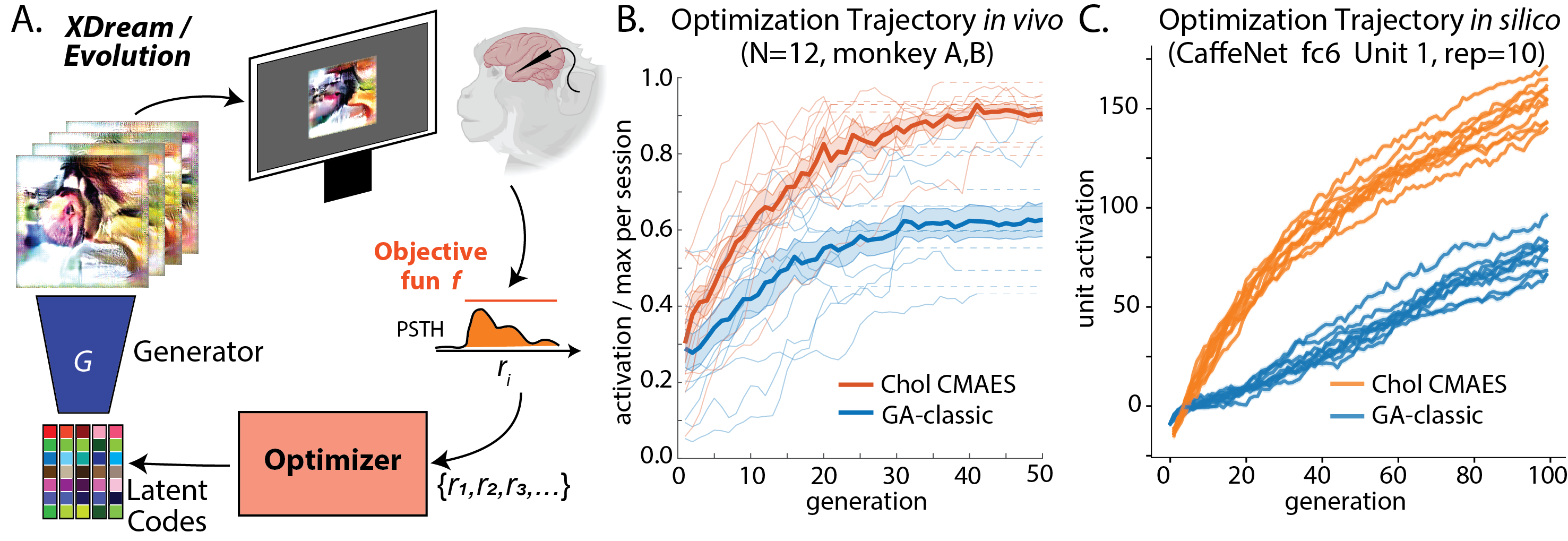}
\vspace{-5pt}
  \caption{Cholsky-CMAES Excelled in Activation Maximization both \textit{in silico} and \textit{in vivo}. \textmd{\textbf{A}. Schematics of the XDream Evolution experiment. \textbf{B.} \textit{in vivo} optimization trajectory  from 12 paired Evolution experiments in two monkeys. Thin curves show the trajectories for individual experiments; shaded thick curves show the mean and standard error (SEM) of trajectories across experiments. Experiments that terminated earlier were extrapolated by constant (dashed line) to match the generation number for mean and SEM calculation. \textbf{C.} \textit{in silico} optimization trajectory comparison for unit 1 in fc6 layer of CaffeNet, mean activation per generation is plotted.}}
  \Description{}\label{fig:teaser_main}\label{fig:GA_CMA_cmp}
\end{teaserfigure}
\maketitle

\section{Introduction}
An essential goal in sensory neuroscience is to define how the neurons respond to natural stimuli and extract useful information to guide behavior. To a first approximation, visual neurons emit high rates of electrical signals to stimuli with certain visual attributes, so their outputs can be interpreted as conveying the presence of such features (e.g. "face neurons"\citep{hubel1962receptive,quiroga2005jenniferAniston}). So to study the visual selectivity of neurons, it is crucial to choose highly activating stimuli.

Traditionally, researchers have used intuition\citep{Gross1994} or limited theoretical frameworks to choose a fixed set of stimuli, i.e., simple images, such as circles and rings for studying lateral geniculate nucleus cells, oriented bars for V1 neurons, hyperbolic gratings for V2 neurons\citep{Hegde2000}, curved shapes for V4 neurons\citep{Pasupathy1999}, or select categories such as faces for inferotemporal cortex neurons\citep{Desimone1984}. The desired property of these stimuli is their ability to drive neuronal activity. But as visual neurons become more selective along the posterior-anterior anatomical axis --- responding to more complex visual attribute combinations --- it becomes more difficult to choose highly activating stimuli.

To tackle this problem, an alternative approach is to use adaptive-search methods to find highly activating stimuli. The idea is to treat neuronal responses as a function of visual stimuli, and to iteratively search for variants that maximize this function. As images evolve under this search, they acquire visual attributes informative of the neuron's intrinsic tuning, independent of bias in human intuition.

\paragraph{Problem Formulation} Formally, neurons can be conceptualized as noisy scalar functions $f$ over the image space $\mathcal I$. One common objective is to maximize the function $f(I)$ \textemdash also known as activation maximization, commonly used for interpreting the coding content of neurons or CNN hidden units \citep{nguyen2016synthPref}: 
\begin{align}
    r=f(I)+\epsilon\;, f:\mathcal{I}\to \mathbb R\\
    I^* = \arg\max_I f(I)
\end{align}
While this approach can be easily generalized to other objectives, we focus on activation maximization since making neurons fire strongly has been the most common goal in sensory neuroscience studies over the decades \citep{hubel1962receptive}. 

In the context of \textit{in vivo} experiments, the process amounts to recording a set of neuronal activations $\{r_i\}$ in response to a set of randomly sampled images $\{I_i\}$ (each image displayed for a short duration, e.g. 100 ms, Fig. \ref{fig:teaser_main}A). In the next step, the optimizers update their state and propose a new set of images. By repeating this process for dozens of iterations, the images begin to acquire visual attributes that drives the neuron's activity highly. Usually, in one of these so-called \textit{Evolution} experiments, the number of image presentations is limited to 1000-3000, taking 20-40 min. This mandates a high sample efficiency of the optimizer. 

For most optimization algorithms, stimuli need to be represented in and generated from a "vector space". We consider a smooth parametrization of images $G:\mathbb R^d\to \mathcal{I},z\mapsto I$ using a lower-dimensional vector space. In this implementation, the mapping from vector space to images is instantiated by an image generative neural network (DeePSim Generative Adversarial Network (GAN) \citep{dosovitskiy2016DeepSiMGAN}), which maps 4096d latent vectors to images. Thus the problem is about searching for images in the latent space such that it maximizes the response of the neuron. 
\begin{displaymath}
    z^*=\arg\max_z \mathbb E [f(G(z))]
\end{displaymath}
This optimization approach has been applied to neurons in visual areas V4 and inferotemporal cortex (IT) \citep{Yamane2008, Rose2021Ponce, Carlson2011Connor,Hung2012Connor,Srinath2021Connor,Ponce2019}. Image search was effectuated by classic genetic algorithms (GAs), acting in the space of parametrized 3D shapes or GANs. 
Though the use of GAs was successful in this domain \citep{Xiao2020XDREAM}, it has not been tested comprehensively against modern optimizers, which motivated us to determine if we could improve performance on this front. 

This problem features a unique set of challenges, for example, search dimensionality is very high (e.g. $d=4096$ in \citep{Xiao2020XDREAM,Ponce2019,Rose2021Ponce}), and the function $f(.)$ must be evaluated with considerable noise in neuronal responses. Further, the total number of function evaluation $N_f$ is highly limited ($N_f<d$), thus the dimensions could not be exhaustively explored. 

In this project, we worked to improve the performance of optimizer in this domain and to extract the underlying principles for designing such optimizer. The main contributions are as follows:
\begin{itemize}
    \item We conducted two \textit{in silico} benchmarks, establishing the better performance of CMA-type optimizers over other optimizers including commonly used genetic algorithms (GA).
    \item We validated the performance increase of the CholeskyCMA algorithm, with a focused contrast to classic GA \textit{in vivo}.
    \item We found that the CMA search trajectories exhibited the spectral signature of high-dimensional guided random walks, preferentially traveling along the top eigen-dimensions of the underlying image manifold. 
    \item We found one reason that CMA succeeded was the decrease of angular step size, thanks to the spherical geometry of image space and the increased vector norms in Evolution.
    \item Guided by image space geometry, we built in these mechanisms to develop a \textit{SphereCMA} algorithm, which outperformed the original CMA \textit{in silico}.
\end{itemize}

\section{Survey of Black Box Optimizers}
Because \textit{in vivo} tests of optimizer performance can be costly and time-consuming, we began by screening a large set of algorithms \textit{in silico} using convolutional neural network (CNN) units as proxies for visual neurons, then tested the top performing algorithms with actual neurons in a focused comparison. 
\subsection{Large Scale \textit{in silico} Survey}
To simulate neuronal tuning function $f$ that an optimizer might encounter in a biological setting, we used units from pre-trained CNNs as models of visual neurons \citep{Xiao2020XDREAM,Lindsay2020CNNasModel}. For the \textit{in vivo} Evolution experiments, we aimed for optimizers that performed well with neurons across visual areas (including V1, V4, IT) and across different levels of signal-to-noise and single-neuron isolation. Thus we designed the benchmark "problem set" to include units from multiple \textit{CNN architectures}, \textit{layers}, and \textit{noise levels}, testing the overall performance for each optimizer. 

We chose AlexNet and a deeper, adversarially trained ResNet (ResNet50-robust) as models of ventral visual cortex. The latter was chosen because it exhibits visual representations similar to the brain (high rank on Brain-Score \citep{Schrimpf2018BrainScore}). For each network, we picked 5 layers of different depths. It has been noted that units from shallow-to-deep layers prefer features of increasing complexity \citep{nguyen2016synthPref}, similar to that in the ventral stream cortical hierarchy \citep{Rose2021Ponce}. For detailed information about these networks and their layers, see Sec. \ref{sec:CNN_details}. 
In the context of \textit{in vivo} recordings, single-presentation neuronal responses are highly noisy \citep{Czanner2015SNRneuron}. To simulate the Poisson-like noisy response $\tilde r$, we added Gaussian noise with zero mean and standard deviation proportional to the raw response $\alpha r$ (ratio $\alpha$ represented the noise level). We tested three noise levels for each objective function: no noise ($\alpha=0$), low noise ($\alpha=0.2$), and high noise ($\alpha=0.5$). 
\begin{equation}
    \tilde r = \max(0,(1+\alpha\epsilon) r),\;\;\epsilon \sim \mathcal{N}(0,1)
\end{equation}
In the first round, we chose 12 gradient-free optimizers as implemented or interfaced by \texttt{nevergrad}\citep{nevergrad}: NGOpt, DE, TwoPointsDE, ES, CMA\citep{Hansen2001CMA-ESorigin}, DiagonalCMA\citep{AkimotoHansen2020DiagonalCMA}, RescaledCMA, SQPCMA, PSO, OnePlusOne, TBPSA, and RandomSearch. Here we compared these algorithms by their default hyper-parameter settings. Note that RandomSearch just sampled random vectors from an isotropic Gaussian distribution, finding the vector with the highest score, which formed the naive baseline (for a short introduction to these algorithms, see \citep{nevergrad}). Each algorithm ran with a budget of 3000 function evaluations and three repetitions per condition. 

\begin{figure}[htp]
  \centering
  \includegraphics[width=0.44\textwidth]{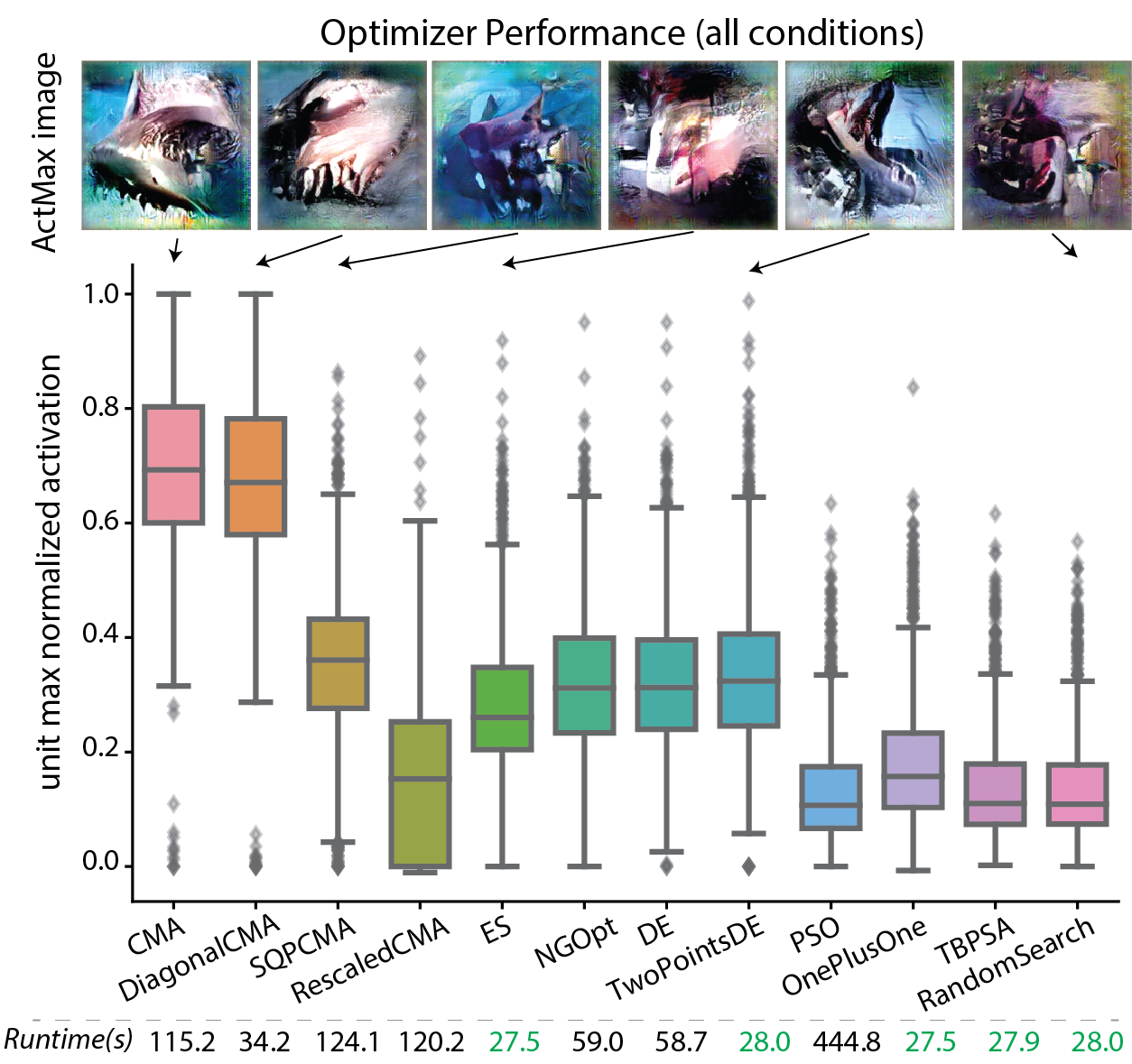}
  \caption{CMAES Excelled in Large Scale Benchmark of Activation Maximization. \textmd{Normalized activation were pooled across all conditions. Mean runtime across all condition were annotated below each optimizer. The activation maximizing image found by the optimizers for an CNN unit (AlexNet-fc8-unit003 tiger shark unit) were showed above the scores.}}
  \Description{}\label{fig:optimizer_benchmark}
  \vspace{-10pt}
\end{figure}

Among the 12 optimizers, we found that Covariance Matrix Adaptation Evolution Strategy (CMA) and DiagonalCMA were the top two algorithms in achieving maximal activation (Fig. \ref{fig:optimizer_benchmark}). Since the upper bound of the activation of a unit in CNN was arbitrary, we divided the raw activations by the empirical maximal activation achieved by that given unit, across all algorithms and repetitions. By this measure, when pooling all conditions, CMAES achieved $0.699\pm0.004$, DiagonalCMA achieved $0.677\pm0.004$, as a reference Random Search baseline achieved $0.139\pm0.002$ (mean $\pm$ sem, $N=1500$, Tab. \ref{tab:ngbenchmark}, Fig. \ref{fig:optimizer_benchmark}. This difference of CMA-driven performance vs. any other optimizer was significant per a two-sample t-test: $t>60$ for all comparisons, except for CMA vs DiagonalCMA, where $t=3.80,p=1.4\times 10^{-4}$). We found the same result held consistently for units across CNN models, layers and noise levels (see comparison in Tab. \ref{tab:ngbenchmark}). 
The optimization score traces and the most activating images found for an example ResNet unit are shown in Fig. \ref{fig:ng_optim_trajectory} and Fig. \ref{fig:nevergrad_proto_cmp}.

As for time efficiency, we measured the total run time taken by optimizing the objective function with a budget of 3000 evaluations\footnote{These optimizers were tested on single core machine with V100 GPU without batch processing of images or activations.}. With units from AlexNet as objective, CMA had a longer runtime of $104.5\pm30.0$ s (mean $\pm$ std); DiagonalCMA accelerated the runtime by roughly five-fold ($23.0\pm4.5$ sec), although at a slightly compromised score (6.1\%). As a reference, the baseline algorithm RandomSearch had an average runtime of $17.6\pm2.2$ sec. Same trends held for ResNet50 units, though the runtime values were generally longer because ResNet50 is deeper (Tab. \ref{tab:ngbenchmark}). In conclusion, we found CMA and DiagonalCMA algorithms were both well-suited for this domain, while DiagonalCMA achieved a good trade-off between speed and performance. 

\subsection{Comparison of CMA-type Algorithm with GA \textit{in silico}} \label{sec:cma_benchmark}
Given the general success of CMA and DiagonalCMA algorithm, we were motivated to test other types of CMA algorithms in the second round, comparing runtime values and achieved activations. 

As described in \citep{Hansen2016Tutorial}, the CMA algorithm maintains and updates a Gaussian distribution $\mathcal N(m,\sigma C)$ in the $\mathbb R^d$ space, with the mean vector $m$ and step size $\sigma$ initialized by the user; the covariance matrix $C$ is initialized as identity matrix $I$. In each step, the algorithm samples a population of codes from this exploration distribution, $z_t^{(i)}\sim \mathcal N(m_t,\sigma_t C_t)$. After evaluating these codes, it updates the mean vector $m_t$ by a weighted combination of the highest ranking codes. Note that $\sigma_t$ controls both the spread of samples in a generation and the average step size of the mean vector update $\|m_{t+1}-m_t\|$. This step size $\sigma_t$ is adapted based on the accumulated path length in the last few steps. Moreover, CMA updates the covariance matrix $C_t$ by a few low-rank matrices to adapt the shape of the exploration distribution. 

In the original CMAES algorithm, after each covariance update, the covariance matrix $C$ needed to be eigen-decomposed, in order to get $C^{-1}$. For a high dimensional space $d=4096$, it is costly to compute this decomposition at each update. Using the diagonal covariance matrix approximation, the inversion step could be simplified from a $\mathcal{O}(n^3)$ operation to $\mathcal{O}(n)$ operation, which makes the DiagonalCMA \citep{AkimotoHansen2020DiagonalCMA} much faster (Fig. \ref{fig:optimizer_benchmark}). This inspired us to use modified covariance update rules to accelerate the optimizer.

We found Cholesky-CMA-ES \citep{Loshchilov2015LM-CMAES} which was proposed as a large scale variant of CMAES. By storing the Cholesky factor $A$ and its inverse of the covariance matrix, it could update these factors directly, without factorizing the covariance matrix at each update. An additional parameter \texttt{A\_update\_freq} could be adjusted to tune the frequency of this update.

We implemented the $(\mu/\mu_w,\lambda)$-Cholesky-CMA-ES algorithm (CholeskyCMA) \citep{Loshchilov2015LM-CMAES} and compared it against the CMA and DiagonalCMA implemented in \texttt{pycma} library \citep{Hansen2014PrincipledDesign,AkimotoHansen2020DiagonalCMA} and the Genetic Algorithm (GA), classically used in this domain \citep{Ponce2019,Srinath2021Connor}. For CholeskyCMA, the hyper-parameters $\sigma_0$ and update frequency of Cholesky factor $A$ were tuned and fixed $\sigma_0=3.0$, \texttt{A\_update\_freq}$=10$. For GA, we used the code and hyperparameters from \citep{Ponce2019}. We allowed for 3000 function evaluations per optimizer, which was 75 generations with a population size of 40. 

We slightly modified the \textit{in silico} benchmark: we chose 7 layers from AlexNet (conv2 to fc8), 10 units from each layer, with 3 noise levels ($\alpha=0.0,0.2,0.5$). Each optimizer were run with 5 repetitions in each condition, totaling 1050 runs. We evaluated the clean score, i.e. the highest noise-free activation of the given unit and the runtime for each optimizer. Here we also used the empirical maximal clean score achieved for each unit to get the normalized clean score and to calculate statistics. 

The results were summarized in Table \ref{tab:CMA_perform_cmp} and Figures \ref{fig:cma_layerwise},\ref{fig:cma_partial_layerwise}. We found that all CMA algorithms outperformed Genetic Algorithm by a large margin: if we pooled all conditions, the mean normalized score of CMA was 166.7\% of that of GA optimizer. Noise deteriorated the score for all optimizers, and there the performance gap between GA and CMA type algorithm was narrowed but persisted: the mean normalized score for CMA was 196.2\% of GA in noise free scenario; 151.9\% for $\alpha=0.2$; 146.4\% for $\alpha=0.5$ (Fig. \ref{fig:cma_layerwise}). The overall performance values of the three CMA-type optimizers were not statistically different, and all surpassed GA. (1-way ANOVA, $F_{2,3147}=1.70,p=0.18$).  

When we examined the performance per layer, we found the relative performance of optimizers had a significant interaction with the source layer  of the unit. DiagonalCMA was more effective than Cholesky and original CMA in the earlier layers, but performed less well in deeper layers (Tab. \ref{tab:CMA_perform_cmp}, Fig. \ref{fig:cma_partial_layerwise}). We tested this interaction with ANOVA on a linear model: \texttt{activation} $\sim$ \texttt{optimizer + noiselevel + layer + layer * optimizer + noiselevel * optimizer}; in which, "optimizer" was modelled as a categorical variable (CholeskyCMA, CMA, DiagonalCMA), "noiselevel" and "layer" were continuous variable. All the factors except \texttt{optimizer} had statistically significant main effect; and interaction term \texttt{layer * optimizer} had $F_{2,3141}=14.98,P=3.34\times 10^{-7}$ (see Tab. \ref{tab:perf_lm_anova_test}). This curious interaction would be interpreted below in the context of different covariance matrix adaption mechanisms (Sec. \ref{sec:dysCovmat}).

As for runtime\footnotemark, DiagonalCMA was still the fastest, with average runtime $6.6\pm0.1$ sec (mean$\pm$ sem, $N=1050$ same below.), while the second fastest optimizer was GA with runtime $16.6\pm 0.2$ sec. In contrast, the classic CMA algorithm taking $97.0\pm 0.8$ was the slowest among the four, while the CholeskyCMA took $42.4\pm0.4$ sec per run. Indeed, updating Cholesky factors made it run faster without reducing performance.
\footnotetext{Here, latent codes were processed in batch of 40 by generator and CNN, thus the same optimizer ran faster than the previous experiment.}

\begin{table}
  \caption{Layer-wise Performance Comparison of CMA-Style Algorithms. \textmd{This table presents mean normalized clean score for each optimizer, averaged across all 10 units and 5 repetitions in given layer under given noise level $\alpha$. For each unit, the clean score was normalized by the highest noise-free activation achieved across optimizers and noise. The last row showed the mean run time ($N=1050$). Abbreviations: Genetic Algorithms (\textbf{GA}), CholeskyCMA (\textbf{Chol}), CMA-ES (\textbf{CMA}) and DiagonalCMA (\textbf{Diag}) as implemented in \texttt{pycma} library, SphereCMA with Exponential decay (\textbf{Sp exp}) or inverse function decay (\textbf{Sp inv}) of step size (see Sec. \ref{sec:sphere_cma}). \red{\textbf{Red}} text showed the best performing optimizer in each condition, the multiple red scores are not statistically different (per $p>0.001$). See Fig. \ref{fig:cma_layerwise} for the distribution of score from all 7 layers.}}
  \vspace{-10pt}
  \begin{tabular}{cl|llll|ll}
\toprule
layer        & $\alpha$ & GA   & Chol                              & CMA                                  & Diag                               & Sp exp                            & Sp inv \\
\midrule
                 & 0.0 & 0.525 & 0.838                 & 0.770                 & 0.823                 & \red{ \textbf{0.901}} & 0.822                 \\
                 & 0.2 & 0.473 & \red{ \textbf{0.654}} & \red{ \textbf{0.626}} & \red{ \textbf{0.685}} & \red{ \textbf{0.659}} & \red{ \textbf{0.637}} \\
\multirow{-3}{*}{\textbf{conv2}} & 0.5 & 0.406 & \red{ \textbf{0.565}} & 0.545                 & \red{ \textbf{0.633}} & 0.548                 & 0.546                 \\
\hline
                 & 0.0 & 0.443 & 0.791                 & 0.793                 & 0.812                 & \red{ \textbf{0.886}} & 0.819                 \\
                 & 0.2 & 0.398 & \red{ \textbf{0.618}} & 0.607                 & \red{ \textbf{0.663}} & 0.603                 & 0.598                 \\
\multirow{-3}{*}{\textbf{conv3}} & 0.5 & 0.345 & \red{ \textbf{0.524}} & \red{ \textbf{0.492}} & \red{ \textbf{0.526}} & \red{ \textbf{0.495}} & \red{ \textbf{0.486}} \\
\hline
                 & 0.0 & 0.378 & 0.758                 & 0.723                 & 0.728                 & \red{ \textbf{0.859}} & 0.754                 \\
                 & 0.2 & 0.358 & 0.554                 & 0.546                 & \red{ \textbf{0.618}} & \red{ \textbf{0.570}} & 0.532                 \\
\multirow{-3}{*}{\textbf{conv4}} & 0.5 & 0.299 & \red{ \textbf{0.446}} & \red{ \textbf{0.435}} & \red{ \textbf{0.468}} & \red{ \textbf{0.440}} & \red{ \textbf{0.420}} \\
\hline
                 & 0.0 & 0.372 & 0.756                 & 0.764                 & 0.760                 & \red{ \textbf{0.913}} & 0.792                 \\
                 & 0.2 & 0.354 & 0.547                 & 0.539                 & \red{ \textbf{0.598}} & 0.542                 & 0.523                 \\
\multirow{-3}{*}{\textbf{conv5}} & 0.5 & 0.298 & \red{ \textbf{0.456}} & \red{ \textbf{0.429}} & \red{ \textbf{0.464}} & \red{ \textbf{0.433}} & 0.409                 \\
\hline
                 & 0.0 & 0.244 & \red{ \textbf{0.571}} & \red{ \textbf{0.577}} & 0.519                 & \red{ \textbf{0.766}} & \red{ \textbf{0.613}} \\
                 & 0.2 & 0.228 & \red{ \textbf{0.361}} & \red{ \textbf{0.370}} & \red{ \textbf{0.429}} & \red{ \textbf{0.404}} & \red{ \textbf{0.399}} \\
\multirow{-3}{*}{\textbf{fc6}}   & 0.5 & 0.201 & \red{ \textbf{0.309}} & \red{ \textbf{0.276}} & \red{ \textbf{0.276}} & \red{ \textbf{0.259}} & \red{ \textbf{0.260}} \\
\hline
                 & 0.0 & 0.320 & \red{ \textbf{0.663}} & \red{ \textbf{0.706}} & 0.659                 & \red{ \textbf{0.830}} & \red{ \textbf{0.662}} \\
                 & 0.2 & 0.295 & \red{ \textbf{0.444}} & \red{ \textbf{0.435}} & \red{ \textbf{0.457}} & \red{ \textbf{0.438}} & \red{ \textbf{0.404}} \\
\multirow{-3}{*}{\textbf{fc7}}   & 0.5 & 0.249 & \red{ \textbf{0.338}} & \red{ \textbf{0.344}} & \red{ \textbf{0.345}} & \red{ \textbf{0.310}} & \red{ \textbf{0.318}} \\
\hline
                 & 0.0 & 0.308 & 0.703                 & 0.733                 & 0.600                 & \red{ \textbf{0.865}} & 0.712                 \\
                 & 0.2 & 0.281 & \red{ \textbf{0.448}} & \red{ \textbf{0.458}} & \red{ \textbf{0.439}} & \red{ \textbf{0.473}} & \red{ \textbf{0.446}} \\
\multirow{-3}{*}{\textbf{fc8}}   & 0.5 & 0.228 & \red{ \textbf{0.328}} & \red{ \textbf{0.319}} & \red{ \textbf{0.324}} & \red{ \textbf{0.316}} & \red{ \textbf{0.305}}\\
\hline
\multicolumn{2}{l|}{\textbf{Average}}	 &	0.333 &	0.556 &	0.547 &	0.563 &	\red{ \textbf{0.596}} &	0.546\\
\hline
\multicolumn{2}{l|}{\textbf{runtime} (s)} & 16.6 & 42.4 & 97.0 & 6.6 & 25.4 & 25.9 \\
\end{tabular}
  \label{tab:CMA_perform_cmp}
  \vspace{-20pt}
\end{table}

\subsection{CMA outperformed GA \textit{in vivo}} \label{sec:cma_ga_invivo}
After the second round of \textit{in silico} benchmarks, we were ready to compete the top candidates with previous state-of-the-art Genetic Algorithm \textit{in vivo}. We chose CholeskyCMA algorithm and compared it against the classic GA, since its speed and performance were both good across noise level and visual hierarchy. For detailed methods of \textit{in vivo} experiments, we refer readers to Sec. \ref{sec:invivo_exp_method}, but briefly, we recorded electrophysiological activity in two animals using chronically implanted arrays, placed within V1, V4 and IT.

We conducted $14$ experiments. Two out of fourteen ($2/14$) experiments did not result in a significant increase in firing rate of the neuron for either optimizer (per criterion $P<0.001$, for t-test between firing rates in first two and last two generation), which we excluded from the further analysis. From the $12$ experiments where at least one optimizer increased the firing rate, we normalized the firing rate of the neuron to the highest generation-averaged firing rate for the CholeskyCMA optimizer (Fig. \ref{fig:GA_CMA_cmp} B). The normalized final generation activation for CholeskyCMA was $0.908\pm0.018$ comparing to $0.628\pm0.045$ for GA (paired t-test, $t=6.69,p=3.4\times10^{-5},df=11$. Raw firing rates for each experiments are shown in Fig. \ref{fig:GA_CMA_cmp_supp}). Thus, the maximal activation of CholeskyCMA surpassed GA by 44\%, which was comparable to the activation gain in the high-noise condition (46.2\%, $\alpha=0.5$) \textit{in silico}. 

From this, we concluded that CholeskyCMA outperformed classic GA algorithm both \textit{in vivo} and \textit{in silico}, becoming the preferred algorithm for conducting activation maximization.


\section{The Analysis of CMA Evolution}
Why did CMA-type algorithms perform so well? Was it the covariance updates, adaptation of step size, or a fortuitous match between the geometric structure of the latent space and the algorithm? We were motivated to find which component contributed to its success. We reasoned the optimizers should work best when they conform to the geometry of the generative image manifold $G$, and the geometry of neuronal tuning function $f$ on the manifold. So we focused on analyzing the geometry of the search trajectory with respect to the geometry of image space \citep{Wang2021Geometry}. When available, we analyzed the \textit{in silico} and \textit{in vivo} evolutions back-to-back to validate the effect. 

\subsection{The "Dysfunction" of Covariance Matrix Adaptation}\label{sec:dysCovmat}%
First, we noticed that in a high-dimensional space, the covariance matrix updates were impaired in the original $(\mu/\mu_W,\lambda)$ CMA or CholeskyCMA algorithm. In the default setting\footnotemark, the covariance learning rates $c_1,c_\mu\propto 1/d^2$, which were exceedingly small at $d=4096$. Thus, the covariance matrix was updated negligibly and could be well approximated by an identity matrix. Recently, this was also pointed out in \citep{AkimotoHansen2020DiagonalCMA} (Sec.4,5) and the authors proposed to increase the learning rate in DiagonalCMA. We tested the effectiveness of this modification.
\footnotetext{See the default setting of $c_1,c_\mu$ in Tab. 2 of \citep{Hansen2014PrincipledDesign}.}

Empirically, we validated this for the original, Cholesky, and Diagonal CMAES algorithms. We quantified this by measuring the condition number of the covariance matrix $\kappa(C)$ and its relative distance to identity matrix $\Delta(C)$. 
\begin{align}
    \kappa(C) = \frac{\lambda_{max}(C)}{\lambda_{min}(C)},\;\;
    \Delta(C) = \frac{\|C- I\|_F^2}{\|C\|_F^2}
\end{align}
We found that the final generation condition number $\kappa(C)$ of the CholeskyCMA was $1.000175\pm0.000004$ (mean $\pm$ std, $N=175$, same below), while the relative distance to identity $\Delta(C)$ was $8.05\pm0.31\times10^{-12}$. In comparison, condition number $\kappa(C)$ for the original CMAES was $1.002739\pm0.000046$, and for the DiagonalCMA, $1.124728\pm0.034774$. The relative distances $\Delta(C)$ to the identity matrix were $7.13\pm0.01\times10^{-8}$ and $1.63\pm0.21\times10^{-4}$ for original CMA and DiagonalCMA algorithms. Though as designed, DiagonalCMA updated the covariance matrix more effectively than the other two CMA algorithms, its final covariance was still quite isotropic. On the other hand, for the original and CholeskyCMA algorithm, we could safely approximate the exploration distribution $p(z^{(i)}_{t+1}|m_t)$ by an isotropic Gaussian scaled by the step size $\sigma_t$
\begin{equation}
    p(z^{(i)}_{t+1}|m_t)\sim \mathcal N(m_t,C)\approx \mathcal N(m_t, \sigma_t I)
\end{equation}
This isotropic exploration throughout the Evolution experiments simplified subsequent analyses of the algorithm.

How is this related to the performance of the algorithm? This relates to the interaction between the unit layer and optimizer, noted above (Sec. \ref{sec:cma_benchmark},Tab. \ref{tab:CMA_perform_cmp},Tab. \ref{tab:perf_lm_anova_test}): the DiagonalCMA outperformed original CMA in earlier layers of CNN but not in deeper layers. It seems the faster update of diagonal covariance matrix was only beneficial for units in earlier layers. The diagonal covariance was designed for a separable, ill-conditioned functional landscape. We noticed that although all units had highly ill-conditioned landscapes, the units in shallower layers had tuning for fewer dimensions than units in deeper layers (Sec. \ref{sec:funct_geomtry}). In other words, the units in earlier layers had a larger invariant space, and a diagonal covariance might suit these landscape better than the more complex ones for deeper units. 

In short, we conclude that the effectiveness of CMA-type algorithm was not in its adaptation of the exploratory distribution shape, and we postulate that it could work with fixed covariance.

\subsection{Evolution Trajectories Showed a Sinusoidal Structure, Characteristic of Random Walks}\label{sec:sinusoidal}
\begin{figure*}[!htp]
  \centering
  \includegraphics[width=0.95\textwidth]{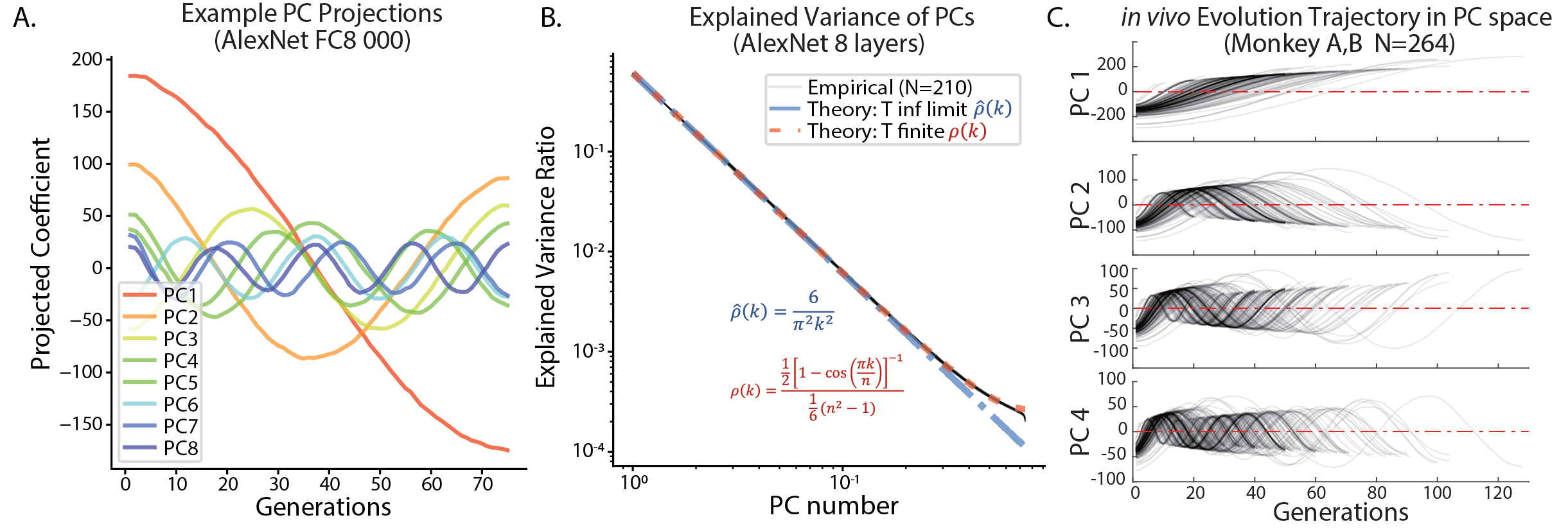}
  \vspace{-10pt}
  \caption{Sinusoidal Geometry of Evolution Trajectory as a Signature of Random Walk. \textmd{\textbf{A.} Mean search trajectory projected onto the first 8 PCs of an example Evolution. \textbf{B.} Explained variance of each PC coincided well with the theoretical value (Eq.\ref{Eq:expvar_eq}) $N=210$ runs plotted, precisely overlapping. \textbf{C.} Mean evolution trajectory projected onto top 4 PCs for $N=264$ Evolution experiments \textit{in vivo}. }}\label{fig:sinusoidal}
  \vspace{-10pt}
  \Description{}
\end{figure*}
Next, we investigated the geometric structure of the Evolution trajectories, through the lens of Principal Component Analysis (PCA), a linear dimension-reduction algorithm. For each experiment, we computed the mean latent vector for each generation $\{\bar{z}_{t}\},t=1...T$, and applied PCA to this $T$-by-$d$ matrix of mean vectors ($T=75$).

We found a pervasive sinusoidal structure to the typical trajectory. When a given trajectory was projected onto the top principal components (PC), the projection resembled cosine waves (Fig. \ref{fig:sinusoidal}A). On the $k$th component, the projected trajectories were well-fit by cosine functions of frequency $k\pi$: $A\cos(k\pi t/T)$ (Fig. \ref{fig:sinusoidal_supp}A). If we allowed the phase and frequency to be fit, $A\cos(2\pi\omega (t/T + \phi))$, the fit $R^2$ was above 0.80 for the top 16 PCs for 95\% of trajectories (Fig. \ref{fig:sinusoidal_supp}B). As a result, projecting the mean trajectory onto the top PCs will result in Lissajous curves (Fig. \ref{fig:LissajousGrid}). Further, we analyzed the PCs of the \textit{in vivo} Evolution experiments ($N=264$), and found the same sinusoidal structure (Fig. \ref{fig:sinusoidal}C). Even for control evolutions driven by noise, this sinusoidal structure persisted (Fig. \ref{fig:sinusoidal_noise_supp}).

This intriguing structure was reminiscent of the PCA of another type of high-dimensional optimization trajectory: that of neural-network parameters during training \citep{lorch2016PCAvisualizingNN}. This structure was later observed and analytically described for high-dimensional random walks or discrete Ornstein-Uhlenbeck (OU) processes guided by potential \citep{antognini2018pcaHighDimRandomWalk}. To test this connection, we examined the explained variance of each principal components (Fig. \ref{fig:sinusoidal}B). We found that the explained variance $\rho(k)$ of the $k$th PC was well approximated by Eq. \ref{Eq:expvar_eq}, which was derived as the theoretical limit for PCA of random walk with $T$ steps in high-dimensional space\footnotemark \citep{antognini2018pcaHighDimRandomWalk}. If we take the step number $T\to\infty$ to infinity in Eq. \ref{Eq:expvar_eq}, the explained variance of $k$th PC scales with $k^{-2}$ (Eq. \ref{Eq:expvar_eq}), showing a fast decay. 
\footnotetext{Our Eq.\ref{Eq:expvar_eq} was adapted from Eq.12 in \citep{antognini2018pcaHighDimRandomWalk}. We found the original equation had the incorrect normalization factor, so we corrected the denominator as it is now. }
\begin{align}
    \rho(k) = \frac{\frac12 [1 - \cos(\frac{\pi k}{T})]^{-1}}{\frac{1}{6}(T^2-1)},\;\;\lim_{T\to\infty}\rho(k) = \frac{6}{\pi^2 k^2}\label{Eq:expvar_eq}
\end{align}
Indeed, both in theory and in our data, the first principal component explained more than $60\%$ variance of mean evolution trajectory, and the top five PC explained close to $90\%$ of the variance (Fig. \ref{fig:sinusoidal} B). In this view, the evolution trajectory of mean latent code $\bar z$ of CMAES could be regarded as a (guided) random walk in high dimensional space, with its main variance residing in a low-dimensional space.

How are evolutionary trajectories related to random walks? In the extreme, when the response is pure noise, the randomly weighted average of the candidates $\{z_t^{(i)}\}$ will be isotropic; thus each step $p(m_{t+1}|m_t)$ is taken isotropically. Intuitively, this is close to a random walk. In our evolutions, the selection of candidates was biased by the visual-attribute tuning of units or neurons, thus their associated trajectories were random walks guided by given potentials. As derived in \citep{antognini2018pcaHighDimRandomWalk}, for random walks guided by quadratic potentials, the structure of trajectory should be dominated by the dimensions with the smallest driving force (curvature). As we will see below (Sec. \ref{sec:hess_align}, Sec. \ref{sec:GAN_geometry},\ref{sec:funct_geomtry}), the latent space had a large portion of dimensions to which the units were not tuned. This may explain why our evolution trajectories looked like random walks via PCA. In summary, we interpreted this as an intriguing and robust geometric property of high-dimensional curves \citep{antognini2018pcaHighDimRandomWalk}, but its full significance remains to be defined. 

\subsection{Evolution Trajectories Preferentially Travel in the Top Hessian Eigenspace}\label{sec:hess_align}

Given that \textit{individual} trajectories had a low-dimensional structure, we asked if there was a common subspace that these Evolution trajectories \textit{collectively} shared. 
We reasoned that each optimization trajectory should lead to a local optimum of the tuning function of each given CNNs unit or cortical neuron, encoding relevant visual features. Though different neurons or units prefer diverse visual attributes of different complexity\citep{Rose2021Ponce}, it was possible that the preferred visual features populated certain subspaces in the latent space of our generator. 

We first collected the \textit{in silico} \textit{Evolution} trajectories of $N=1050$ runs, across all conditions, and represented each trajectory by the mean code from the final generation, i.e. the evolution direction $\zeta_j$. We shuffled the entries of each vector $\{\zeta_j\},j=1...N$ to form a control collection of evolution directions $\{\zeta_{shfl}\}$, preserving their vector norms. We found that the collection of observed \textit{Evolution} directions was lower-dimensional than its shuffled counterpart: by PCA, the explained variance of top 7 PCs was larger than the shuffled counterpart ($p<0.005$  comparing to 500 shuffles, Fig. \ref{fig:hessian_align_supp}).

\begin{figure}[htp]
  \centering
  \includegraphics[width=0.48\textwidth]{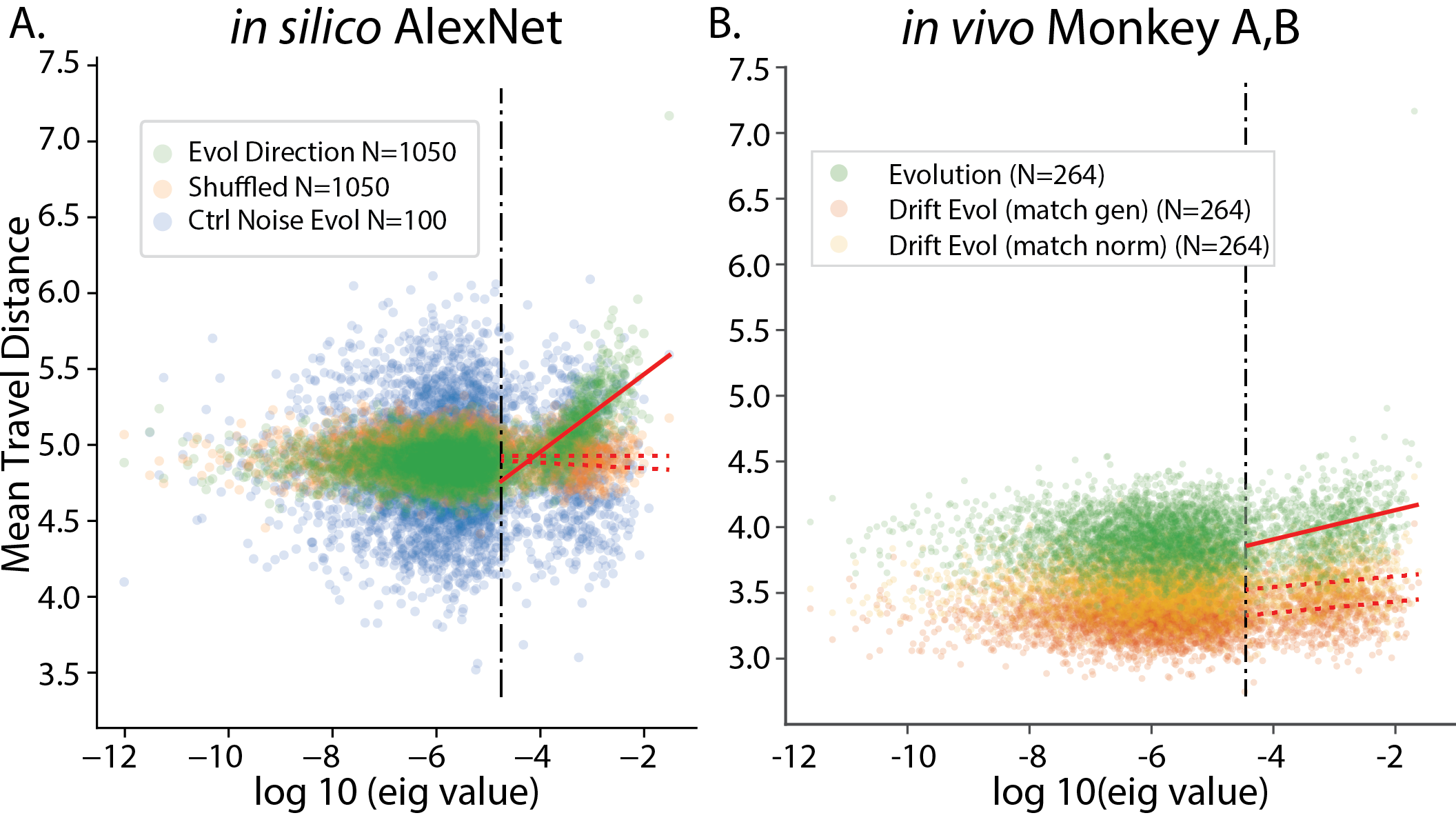}
  \vspace{-15pt}
  \caption{Evolution Directions Preferably Aligned with the Top Hessian Eigenspace.  \textmd{Mean distance traveled along each eigenvector, plotted against the log eigenvalue. \textbf{A.} \textit{In silico} evolutions, shuffled code and evolution driven by noise. The cutoff line (at 800th eigenvalue) plotted in dashed line. Regression lines plotted for the top 800 points for each condition. \textbf{B.} Same as \textbf{A} but for \textit{in vivo} evolutions, and \textit{in silico} noise-driven evolutions matching the generation number or norm of each \textit{in vivo} evolution. }} 
  
  \Description{}\label{fig:hessian_align}
  \vspace{-10pt}
\end{figure}

Next, we examined the relationship between the collection of trajectories and the global geometry of the space (Sec. \ref{sec:GAN_geometry}). Previously, we reported that the latent space of the generator exhibits a highly anisotropic geometry, as quantified by the Riemannian metric tensor $H$ of this image manifold \citep{Wang2021Geometry}. The bilinear form $v^THv$ defined by this Riemannian metric tensor represented the degree of image change when moving in direction $v$. Consequently, moving in top eigen-directions changes the generated image much more than other directions, while moving in the bottom eigenspace scarcely affects the image (Fig. \ref{fig:supp_GANgeometry}B,C). Moreover, this structure is homogeneous in the space, thus similar directions will cause rapid image change at different positions in the latent space. Thus, there exist \textit{global dimensions} that affect the image a lot, and dimensions that do not. We hypothesized that the optimization trajectories might preferentially travel in some part of this eigenspace. 

We used the averaged metric tensor $H$ from Wang and Ponce (2021, \citep{Wang2021Geometry}). Its eigenvectors $U=[u_1,u_2...u_d]$ with eigenvalues $\lambda_k,k=1..,d$ became our reference frame for analyzing trajectories. We projected the collection of evolution directions $\{\zeta_j\}$ onto this basis, and examined the mean projection amplitude on each eigenvector $\frac{1}{P}\sum_i^P |u_k^T\zeta_i|$ as a function of eigenvalue $\lambda_k$ (Fig. \ref{fig:hessian_align}A). Since for \textit{in silico} experiments, the initial generation was the origin, this quantity was the average distance travelled along the eigenvector across runs. We expected the trajectories to travel further for eigen-dimensions where more CNN units exhibit "gradient". 

We applied the Kolmogorov–Smirnov test to determine if the distribution of projection coefficient in the top and bottom eigen-dimensions were different. We found that the projection coefficients onto the top $800$ eigen-dimensions as a distribution were significantly different from those onto bottom dimensions (KS statistics $0.223, P=9.3\times 10^{-18}$, Fig. \ref{fig:hessian_align}A). Further, within the top $800$ PC dimensions, the mean traveled distance strongly correlated with the log eigenvalues (Pearson correlation $0.738,p=2.8\times10^{-138}$, Fig. \ref{fig:hessian_align}A), i.e. the trajectories tended to travel further for dimensions with larger eigenvalues.

As for \textit{in vivo} evolutions, we examined the collection of evolution trajectories ($N=264$ from 2 monkeys, \citep{Rose2021Ponce}) and projected them onto the Hessian eigenframe as above. We found that they preferentially aligned with the top eigenspace than the lower eigenspace (Fig. \ref{fig:hessian_align}B): the distance traveled in the top $800$ eigenspace correlated with the eigenvalues (Pearson correlation $0.319,p=1.2\times 10^{-15}$). Since \textit{in vivo} evolutions used a set of initial codes $\{z^{(i)}_0\}$ instead of the origin, we used noise-driven evolutions starting from these initial codes as control. As a baseline, the noise-driven evolutions had a lower correlation between the distance traveled and the eigenvalues ($r=0.198$ for the noise evolution with matching generations; $r=0.223$ for that with matching code norm). Our results were robust to the choice of cutoff dimension (600-1000). 

In conclusion, this result showed us that the evolution trajectories traveled further in the the top eigenspace, and the average distance traveled was positively correlated with the eigenvalue. How do we interpret this effect? Since the top eigen-dimensions change the images more perceptibly, the tuning functions of neurons and CNN units were more likely to exhibit a "gradient" in such subspace. As the lower eigenspace did not induce perceptible changes, those dimensions barely affected the activations of units or neurons. Thus, no signal could guide the search in lower eigenspace, inducing dynamics similar to pure diffusion. As the visually tuned units could exhibit gradient in the top eigenspace, the search would be a diffusion with a driving force, allowing the optimizers to travel farther in the top eigenspace. 

\subsection{The Spherical Geometry of Latent Space Facilitated Convergence}\label{sec:norm_converg}
\begin{figure}[htp]
  \centering
\vspace{-10pt}
  \includegraphics[width=0.49\textwidth]{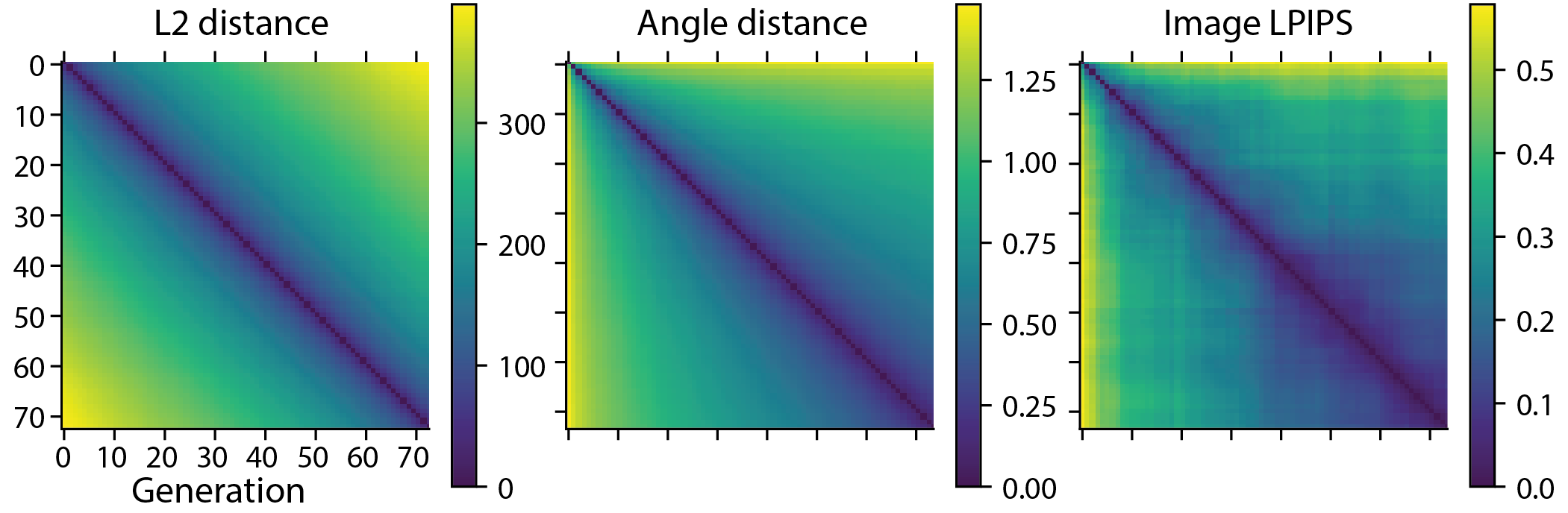}
  \caption{Angular Distance as a Better Proxy for Image Distance. \textmd{We show L2 and angular distance between pairs of mean codes $\bar z_t$ across 75 generations, and perceptual distance (LPIPS) between pairs of images generated from mean codes. }}
  \Description{}\label{fig:L2_ang_LPIPScmp}
\vspace{-8pt}
\end{figure}
Finally, we noticed one geometric structure that facilitated the convergence of the CMA algorithm. For the DeePSim generator\citep{dosovitskiy2016DeepSiMGAN} with DCGAN architecture\citep{radford2015DCGAN}, the mapping $G$ was relatively linear. Namely, changing the scale of the input $z$ mainly affected the contrast of the generated image $G(z)$ (Fig. \ref{fig:supp_GANgeometry}). Thus, when the base vector $z_0$ has a larger norm, travelling the same \textit{euclidean distance} $\Delta z$ will induce a smaller perceptual change; in contrast, traveling the same angular deviation $\Delta \theta$ will result in a similar pattern change regardless of the norm of the base vector $z_0$ (Fig. \ref{fig:example_ang_eucl}). We reasoned that the \textit{angular distance} in latent space would be a better proxy to the \textit{perceptual distance} across generated images.

\begin{figure}[!htp]
  \centering
  \includegraphics[width=0.31\textwidth]{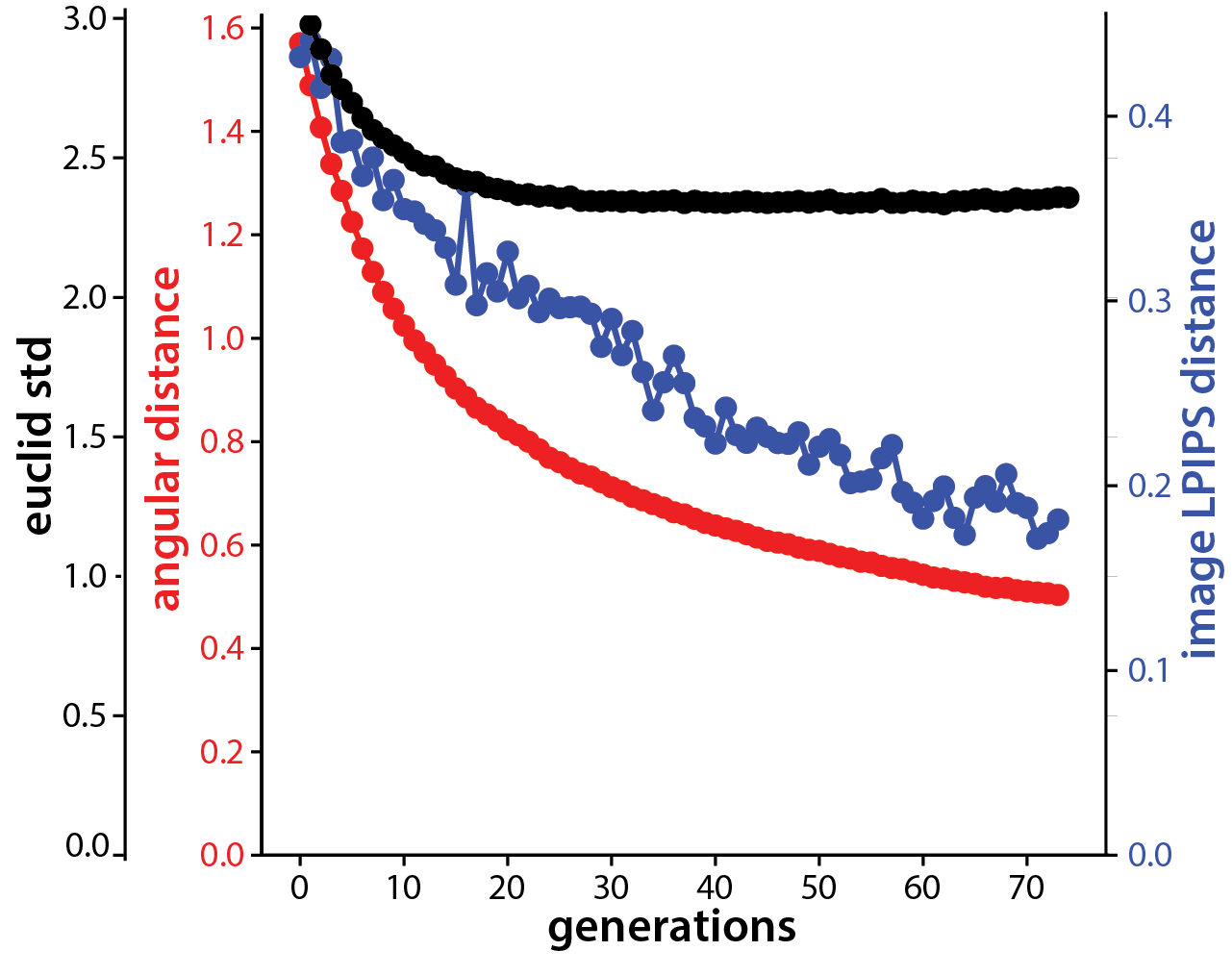}
\vspace{-10pt}
  \caption{Perceptual Variability Better Modelled by Angular Distance. \textmd{We plot the average angular distance of latent codes within each generation; Euclidean standard deviation $\tilde \sigma_t$; and the mean LPIPS image distance within each generation. }}
  \Description{}\label{fig:L2_ang_LPIPScmp_2}
\vspace{-15pt}
\end{figure}
We validated this using the perceptual dissimilarity metric \textit{Learned Perceptual Image Patch Similarity} (LPIPS, \citep{zhang2018LPIPS}). We measured image variability in each generation using the mean LPIPS distance between all pairs. We measured the code variability by the standard deviation (std) of latent codes $\tilde\sigma_t$ averaged across $d$ dimensions, which was an estimate of the step size $\sigma_t$ and was strongly correlated with the mean L2 distance among codes. We also measured the mean angular distance between all pairs of codes in each generation. We found that the std of latent codes $\tilde \sigma_t$ decreased from $\tilde \sigma_0=3.0$ (first generation) to $\tilde \sigma_T=2.31\pm0.04$ ($N=1050$), and hit the floor at around 25 generations (Fig. \ref{fig:L2_ang_LPIPScmp_2}). In contrast, the perceptual variability kept decreasing till the end, which was better approximated by the trend of angular variability within each generation (from 1.57 rad to 0.50 rad, Fig. \ref{fig:L2_ang_LPIPScmp_2}). This discrepancy of Euclidean and angular distance was mediated by the increasing vector norm during evolution: we found that the squared norm of the mean latent codes $\bar z_t$ scaled linearly with generation $\|\bar z_t\|^2\propto t$ ($r^2=0.9985\pm0.008$, $N=1050$, Fig. \ref{fig:sinusoidal_noise_supp}C), which is a classic property of a random walk. As a combined effect of the increasing vector norm and the decreasing step size $\sigma_t$, the angular variability within a generation decreased towards the end. 
Similarly, we found that across each given trajectory, angular distance was also a better proxy for the perceptual distance between images than L2 distance (Fig. \ref{fig:L2_ang_LPIPScmp}). The mean-code images from late generations were more similar to each other than the ones from initial generations, as predicted by the angular distance between the mean vectors --- but not L2 distance.

For a visual neuron or a CNN unit, the perceptual distances across images matter more. Thus, when designing the new optimizer, we reasoned it should control the angular step size as a proxy to perceptual distance, instead of the Euclidean distance ($\sigma$) (Sec. \ref{sec:sphere_cma}). 

\section{Proposed Improvement: SphereCMA}\label{sec:sphere_cma}
Finally, we proposed an optimizer incorporating the findings above. We would test whether this new optimizer could perform as well as the original CMA algorithms. 

This optimizer is designed to operate on a hypersphere, thus it was called SphereCMA (Alg. \ref{alg:sph_CMA}). This design was guided by the training of the image generator $G$. Many generative models (GAN) were trained to map an isotropic distribution of latent codes $p(z)$ (e.g. Gaussian \citep{karras2019styleGAN,dosovitskiy2016DeepSiMGAN} or truncated Gaussian \citep{brock2018BigGAN}) to a distribution of natural images. Thus, the generator $G$ has only learned to map latent codes $z$ sampled from this distribution. In a high-dimensional $d$ space, the density of standard normal distribution concentrates around a thin spherical shell with radius $R \approx \sqrt{d}$. Due to this, latent codes should be sampled from this hypersphere to obtain natural images. Thus, we enforced the optimizer to search on this sphere.

\begin{figure}[htp]
\vspace{-5pt}
  \centering
  \includegraphics[width=0.34\textwidth]{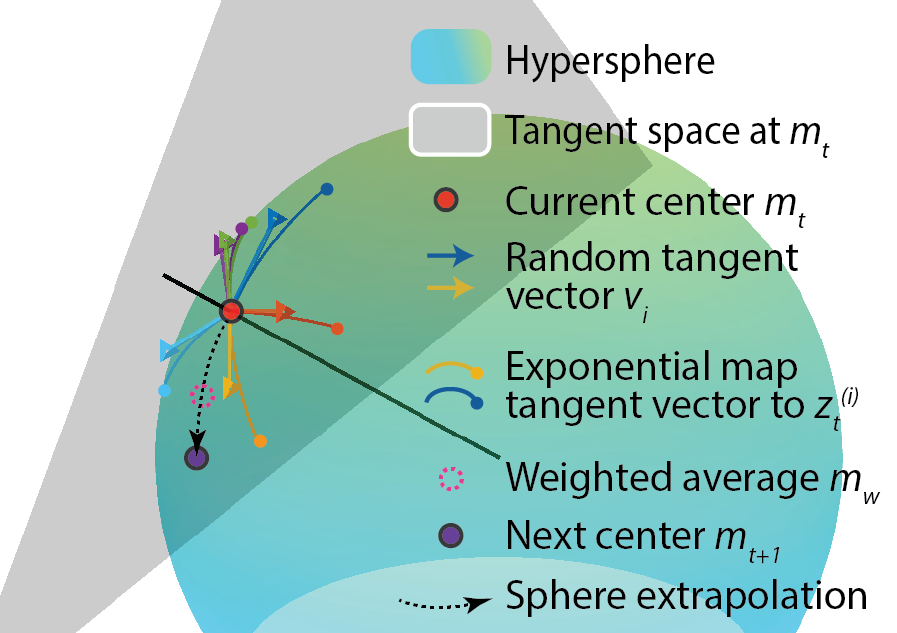}
\vspace{-10pt}
  \caption{Schematics of SphereCMA Update Procedure}
  \Description{}\label{fig:sphereCMA_schmatics}
\vspace{-15pt}
\end{figure}

To design it, we leveraged the principles we learned from CMA-ES and the geometry of the latent space. First, since higher contrast images usually stimulate neurons or CNN units better, we set the spherical optimizer to operate at a norm comparable to that achieved by the end of a successful evolution experiment ($R=300$), ensuring a high contrast from the start. Second, since the covariance matrix update did not contribute to the success of the CMA-ES algorithm in this application (Sec. \ref{sec:dysCovmat}), we kept the algorithm exploring in an isotropic manner, without covariance update. To sample codes on the sphere, we used a trick from differential geometry, sampling vectors $v_i$ isotropically in the tangent space of the mean vector $m_t$, and then mapping these tangent vectors onto the sphere as new samples $z^{(i)}_{t+1}$ (Fig. \ref{fig:sphereCMA_schmatics}). 
Third, since the angular distance better approximated the perceptual change in the image (Sec. \ref{sec:norm_converg}), we controlled the perceptual variability in each generation by sampling codes with a fixed angular distance (the step size $\mu$) to the mean vector. Fourth and final, as the latent vector norm was fixed, the SphereCMA lacked the step size shrinking mechanism endowed by increasing code norm (Sec. \ref{sec:norm_converg}). We built in an angular step size decay function $\mu_{dec}(t)$ modelling that in Fig. \ref{fig:L2_ang_LPIPScmp_2}, to help convergence. 

\begin{algorithm}
\caption{SphereCMA}\label{alg:sph_CMA}
\begin{algorithmic} 
\Require Objective $f(.)$, space dimension $d$, radius $R$, 
population size $B$, step size decay function $\mu_{dec}(.)$, learning ratio $lr=1.5$
\Require Subroutines \texttt{ExpMap} (Alg.\ref{alg:expmap}), \texttt{SphereExtrapolate} (Alg.\ref{alg:SphExtrapolation}) \texttt{RankWeight} (Alg.\ref{alg:rankweight})

\State Isotropically sample $B$ vectors $z^{(i)}_{0}$ with length $R$ in $\mathbb R^d$ 
\For{$t=0:$ \texttt{maxstep}}
\State Evaluate objective function $f$ for samples $r_i=f(z^{(i)}_{t})$
\State Compute weights $w$ based on the rank of scores $r$. 
\State Weighted average the samples $m_w \gets \sum_i w_i z^{(i)}_{t}/\sum_i w_i$ 
\State Normalize $m_w$ to norm $R$, $m_w\gets Rm_w/\|m_w\|$ 
\State Fetch the current center vector $m_{t}\gets z^{(0)}_{t}$ 
\State Calculate the new center vector $m_{t+1}$ by spherical extrapolation along the arc from $m_{t}$ to $m_w$ by $lr$.
\State Sample $B$ isotropic random vectors, e.g. $u_i\sim \mathcal N(0,I^d)$. 
\State Project $u_i$ to the tangent space of new center $m_{t+1}$. $v_i\gets u_i - m_{t+1}m_{t+1}^Tu_i/\|m_{t+1}\|^2$ 
\State Get current angular step size $\mu \gets \mu_{dec}(t)$        
\State Get new samples $z_{t+1}^{(i)}$ by exponential map from the center $m_{t+1}$ along the tangent vectors $v_i$ with angle $\mu$ 
\State Add the new center vector to the population $z_{t+1}^{(0)} \gets m_{t+1}$
\EndFor
\end{algorithmic}
\end{algorithm}

We tested this algorithm against other CMA-style algorithms in the same setup as in Sec. \ref{sec:cma_benchmark}. Specifically, we tested two versions with different step-size-decay function $\mu_{dec}$, exponential (\textbf{Sp Exp}) and inverse function (\textbf{Sp Inv}). The change of angular step size through generations is shown in Fig. \ref{fig:Sph_step_tune}, the parameters for the decay function were tuned by Bayesian optimization and fixed. 

We found that, SphereCMA with exponential decay (Sp Exp) outperformed all other CMA algorithms when pooled across all noise levels and layers (Tab. \ref{tab:CMA_perform_cmp}, two-sample t-tests, $t_{2098}=4.03,p=5.7\times 10^{-5}$ for CholeskyCMA, $t_{2098}=4.92,p=9.1\times 10^{-7}$ for original CMA, $t_{2098}=3.30,p=1.0\times 10^{-3}$ for DiagonalCMA, $N=1050$). Compared to CholeskyCMA, SphereCMA surpassed its achieved activation by around 7\%. 
Interestingly, when we compared the performance per noise level, we found that SphereCMA-Exp was the best performing algorithm under the noise-free condition ($\alpha=0$) across the seven layers (two-sample t-test, $t_{698}>9, p<1\times10^{-20}$ for all of CholeskyCMA, original CMA and DiagonalCMA; runs were pooled across 7 layers, $N=350$, same below). But in noisier conditions, it performed comparably or sometimes worse than other CMA optimizers: at $\alpha=0.2,0.5$ all comparisons with other optimizers were not significant except for SphereCMA-Exp vs DiagonalCMA at $\alpha=0.5$ ($t_{698}=-3.08, p=2.2\times10^{-3}$). For trajectory comparison, see Fig. \ref{fig:cmagasphere_optim_trajectory}.

Overall, this result confirmed that the principles extracted from analyzing CMA-style algorithms were correct and crucial for their performance. By building these essential structures into our spherical optimizer, we could replicate and even improve the performance of CMA-ES algorithms. Finally, the interaction between the performance of SphereCMA-Exp and the noise level, provided us a future direction to find the source of noise resilience in the original CMA algorithm and improve on the SphereCMA algorithm. 

\section{Discussion}
In this study, we addressed a problem common to both machine learning and visual neuroscience -- the problem of defining the visual attributes learned by hidden units/neurons across the processing hierarchy. 
Neurons and hidden units are highly activated when the incoming visual signal matches their encoded attributes, so one guiding principle for defining their encoded information is to search for stimuli that lead to high activations. Since the brain does not lend itself to gradient descent, it is necessary to use evolutionary algorithms to maximize activations of visual neurons \textit{in vivo}.
Here, we identified a class of optimizers (CMA) that work well for this application and analyzed why they perform so well, and finally developed a faster and better optimizer (SphereCMA) based on these analysis. 
Here are some lessons we learned in this exploration. First, screening with comprehensive \textit{in silico} benchmarks accelerates the algorithm development \textit{in vivo}. 
Secondly, geometry of latent space matters. As a general message, when developing optimizers searching in the latent space of some generative models, we should pay attention to the distance structure i.e. geometry of the generated samples (e.g. images) instead of the latent space. Optimizers leveraging space geometry shall perform better. We hope our workflow can help the optimizer design in other domains, e.g. search for optimal stimuli in other sensory modalities and drug discovery in molecular space. 

\begin{acks}
We appreciate Mary Burkemper in her assistance in monkey managing and data collection. We thank Yunyi Shen, Kaining Zhang for reading the manuscript and providing valuable feedbacks. We are grateful to the RIS cluster and staff in Washington University in St Louis for their GPU resources and technical support.

B.W. is funded by the McDonnell Center for Systems Neuroscience in Washington University (pre-doctoral fellowship to B.W.). Additional research funding by the David and Lucile Packard Foundation and the E. Matilda Ziegler Foundation for the Blind (C.R.P.). 
\end{acks}

\bibliographystyle{ACM-Reference-Format}
\bibliography{main}

\newpage
\appendix
\counterwithin{figure}{section}
\counterwithin{table}{section}

\section{Detailed Methods}
\subsection{Details of Pre-trained CNN Models}\label{sec:CNN_details}
In this work, we used units from two pretrained CNN models, AlexNet and ResNet50-robust. Both were pre-trained for object recognition on ImageNet\citep{deng2009imagenet2012}. 

\textbf{AlexNet}: The architecture and weights were fetched from torchvision.modelzoo. As for layers, we used the activations from conv2-fc8 layer, the activation were non-negative (post ReLU rectification) except for fc8, which we used the logits. 

\textbf{ResNet50-robust}: The architecture were defined in torchvision.modelzoo, with the weights fetched from \url{https://github.com/MadryLab/robustness}. We used the version with adversarial training setting: $L_\infty$ norm with strength $8/255$. It has been noticed that the adversarially trained network exhibited more perceptually aligned gradients, the direct gradient descent from the units could generate impressive feature visualizations. We used this model in our benchmark since the achieved activations were harder to be affected by "adversarial" patterns generated by the generator. For layers, we used the output of last Bottleneck block from layer1-4, namely {layer1.Bottleneck2}, {layer2.Bottleneck3}, {layer3.Bottleneck5}, {layer4.Bottleneck2}, and the {linear} layer before softmax.


\begin{table}[htp]
\caption{Statistical Test of Factors Affecting of CMA-type Algorithm Performance \textmd{Model formula in R style: \texttt{score} $\sim$ \texttt{C(optimizer) + noiselevel + layernum + C(optimizer):layernum + C(optimizer):noiselevel}. The factor \texttt{optimizer} had three categorical levels: CholeskyCMA, CMA, DiagonalCMA, \texttt{noiselevel} had three continuous values: $0.0,0.2,0.5$, \texttt{layernum} had seven continuous values, 2 to 8 encoding conv2 to fc8. Data were subset of the benchmark dataset 2, Tab. \ref{tab:CMA_perform_cmp}.}}
\vspace{-10pt}
\begin{tabular}{l|rrrl}
\toprule
                       & sum sq & df   &       F  &      Pr(>F)  \\
\hline
optimizer              &   0.14 &  2   &    3.40  &  $3.36\times10^{-02}$  \\
layernum               &  19.14 &  1   &  953.56  & $4.39\times10^{-183}$  \\
noiselevel             &  43.56 &  1   & 2170.60  &  $0.0$                 \\
optimizer:layernum     &   0.60 &  2   &   14.98  &  $3.33\times10^{-07}$  \\
optimizer:noiselevel   &   0.19 &  2   &    4.66  &  $9.58\times10^{-03}$  \\
\hline
Residual               &  63.03 & 3141 &          &                        \\
\toprule
\end{tabular}\label{tab:perf_lm_anova_test}
\end{table}

\begin{table*}[t]
\caption{Performance Comparison of Nevergrad Optimizers per Layer and Noise-level. 
\textmd{CMA and DiagonalCMA excelled at all noise level and layers than other algorithms. The performance gaps were larger for deeper layers and higher noise level.  Abbreviations: \textbf{RN50}: ResNet50-robust; \textbf{L1-Btn2}: layer1-bottleneck2. Name of optimizers: \textbf{Diag}: DiagonalCMA; \textbf{SQP}: SQPCMA; \textbf{Rescaled}: RescaledCMA; \textbf{NG}: NGopt; \textbf{2pDE}: TwoPointsDE; \textbf{1+1}: OnePlusOne; \textbf{Rand}: RandomSearch. These full names could be used to fetch optimizers from the registry of nevergrad package by \texttt {ng.optimizers.registry[optimname]}. Due to the limitation of interface of nevergrad, we didn't record the clean score before adding noise. This should make the score distribution noisier, but will not change the rank of the algorithms.}}
    
\label{tab:ngbenchmark}  
\begin{tabular}{cl|llll|lllllll|l}
\toprule
net-layer                              & $\alpha$ & CMA                  & Diag          & SQP & Rescaled & ES & NG & DE    & 2pDE   & PSO   & 1+1 & TBPSA & Rand \\
\hline
                                            & 0.0 & \red{\textbf{0.478}} & \red{\textbf{0.473}} & 0.345  & 0.218   & 0.325 & 0.359 & 0.347 & 0.358 & 0.222 & 0.237 & 0.209 & 0.213\\
                                            & 0.2 & \red{\textbf{0.599}} & \red{\textbf{0.582}} & 0.426  & 0.240   & 0.416 & 0.425 & 0.417 & 0.428 & 0.274 & 0.322 & 0.255 & 0.252\\
\multirow{-3}{*}{\textbf{RN50-L1-Btn2}}     & 0.5 & \red{\textbf{0.809}} & \red{\textbf{0.801}} & 0.558  & 0.326   & 0.572 & 0.555 & 0.575 & 0.607 & 0.376 & 0.443 & 0.367 & 0.361\\
\hline
                                            & 0.0 & \red{\textbf{0.545}} & \red{\textbf{0.534}} & 0.346  & 0.318   & 0.331 & 0.351 & 0.360 & 0.370 & 0.197 & 0.203 & 0.208 & 0.208\\
                                            & 0.2 & \red{\textbf{0.637}} & \red{\textbf{0.644}} & 0.401  & 0.326   & 0.410 & 0.411 & 0.392 & 0.449 & 0.220 & 0.281 & 0.249 & 0.250\\
\multirow{-3}{*}{\textbf{RN50-L2-Btn3}}     & 0.5 & \red{\textbf{0.831}} & \red{\textbf{0.825}} & 0.529  & 0.442   & 0.565 & 0.574 & 0.520 & 0.606 & 0.303 & 0.409 & 0.335 & 0.338\\
\hline
                                            & 0.0 & \red{\textbf{0.578}} & \red{\textbf{0.567}} & 0.344  & 0.174   & 0.230 & 0.283 & 0.279 & 0.301 & 0.111 & 0.121 & 0.131 & 0.128\\
                                            & 0.2 & \red{\textbf{0.619}} & \red{\textbf{0.600}} & 0.357  & 0.173   & 0.270 & 0.311 & 0.308 & 0.325 & 0.123 & 0.176 & 0.142 & 0.138\\
\multirow{-3}{*}{\textbf{RN50-L3-Btn5}}     & 0.5 & \red{\textbf{0.846}} & \red{\textbf{0.787}} & 0.458  & 0.203   & 0.370 & 0.422 & 0.411 & 0.411 & 0.164 & 0.234 & 0.206 & 0.205\\
\hline
                                            & 0.0 & \red{\textbf{0.624}} & \red{\textbf{0.631}} & 0.259  & 0.139   & 0.188 & 0.236 & 0.227 & 0.243 & 0.096 & 0.104 & 0.090 & 0.095\\
                                            & 0.2 & \red{\textbf{0.667}} & \red{\textbf{0.656}} & 0.272  & 0.124   & 0.217 & 0.238 & 0.243 & 0.258 & 0.105 & 0.134 & 0.106 & 0.106\\
\multirow{-3}{*}{\textbf{RN50-L4-Btn2}}     & 0.5 & \red{\textbf{0.724}} & \red{\textbf{0.740}} & 0.267  & 0.130   & 0.279 & 0.294 & 0.316 & 0.335 & 0.137 & 0.192 & 0.149 & 0.141\\
\hline
                                            & 0.0 & \red{\textbf{0.721}} & \red{\textbf{0.651}} & 0.311  & 0.225   & 0.190 & 0.214 & 0.238 & 0.256 & 0.090 & 0.096 & 0.093 & 0.095\\
                                            & 0.2 & \red{\textbf{0.664}} & \red{\textbf{0.678}} & 0.336  & 0.151   & 0.218 & 0.234 & 0.242 & 0.252 & 0.098 & 0.132 & 0.108 & 0.107\\
\multirow{-3}{*}{\textbf{RN50-fc}}          & 0.5 & \red{\textbf{0.760}} & \red{\textbf{0.753}} & 0.351  & 0.163   & 0.292 & 0.286 & 0.304 & 0.324 & 0.135 & 0.171 & 0.139 & 0.144\\
\midrule
                                            & 0.0 & \red{\textbf{0.592}} & \red{\textbf{0.585}} & 0.254  & 0.100   & 0.236 & 0.317 & 0.324 & 0.338 & 0.087 & 0.135 & 0.075 & 0.081\\
                                            & 0.2 & \red{\textbf{0.643}} & \red{\textbf{0.657}} & 0.278  & 0.080   & 0.284 & 0.387 & 0.356 & 0.371 & 0.096 & 0.181 & 0.086 & 0.087\\
\multirow{-3}{*}{\textbf{Alex-conv2}}       & 0.5 & \red{\textbf{0.792}} & \red{\textbf{0.751}} & 0.331  & 0.093   & 0.374 & 0.421 & 0.438 & 0.423 & 0.106 & 0.242 & 0.107 & 0.112\\
\hline
                                            & 0.0 & \red{\textbf{0.658}} & \red{\textbf{0.657}} & 0.327  & 0.125   & 0.214 & 0.306 & 0.310 & 0.298 & 0.083 & 0.092 & 0.078 & 0.075\\
                                            & 0.2 & \red{\textbf{0.689}} & \red{\textbf{0.697}} & 0.296  & 0.100   & 0.245 & 0.298 & 0.316 & 0.307 & 0.083 & 0.140 & 0.084 & 0.080\\
\multirow{-3}{*}{\textbf{Alex-conv3}}       & 0.5 & \red{\textbf{0.822}} & \red{\textbf{0.762}} & 0.314  & 0.112   & 0.336 & 0.376 & 0.370 & 0.367 & 0.110 & 0.184 & 0.106 & 0.106\\
\hline
                                            & 0.0 & \red{\textbf{0.710}} & \red{\textbf{0.661}} & 0.342  & 0.014   & 0.186 & 0.259 & 0.271 & 0.267 & 0.059 & 0.073 & 0.067 & 0.064\\
                                            & 0.2 & \red{\textbf{0.676}} & \red{\textbf{0.656}} & 0.362  & 0.012   & 0.227 & 0.286 & 0.262 & 0.267 & 0.068 & 0.113 & 0.072 & 0.072\\
\multirow{-3}{*}{\textbf{Alex-conv5}}       & 0.5 & \red{\textbf{0.798}} & \red{\textbf{0.764}} & 0.438  & 0.018   & 0.295 & 0.346 & 0.343 & 0.335 & 0.090 & 0.163 & 0.089 & 0.094\\
\hline
                                            & 0.0 & \red{\textbf{0.824}} & \red{\textbf{0.722}} & 0.318  & 0.117   & 0.202 & 0.259 & 0.277 & 0.276 & 0.086 & 0.121 & 0.084 & 0.082\\
                                            & 0.2 & \red{\textbf{0.688}} & \red{\textbf{0.617}} & 0.324  & 0.080   & 0.242 & 0.273 & 0.281 & 0.277 & 0.097 & 0.159 & 0.093 & 0.091\\
\multirow{-3}{*}{\textbf{Alex-fc7}}         & 0.5 & \red{\textbf{0.749}} & \red{\textbf{0.739}} & 0.354  & 0.098   & 0.327 & 0.381 & 0.351 & 0.377 & 0.136 & 0.219 & 0.126 & 0.122\\
\hline
                                            & 0.0 & \red{\textbf{0.814}} & \red{\textbf{0.768}} & 0.334  & 0.199   & 0.173 & 0.224 & 0.223 & 0.228 & 0.069 & 0.089 & 0.089 & 0.087\\
                                            & 0.2 & \red{\textbf{0.680}} & \red{\textbf{0.672}} & 0.351  & 0.137   & 0.205 & 0.221 & 0.237 & 0.243 & 0.076 & 0.117 & 0.098 & 0.099\\
\multirow{-3}{*}{\textbf{Alex-fc8}}         & 0.5 & \red{\textbf{0.744}} & \red{\textbf{0.695}} & 0.382  & 0.127   & 0.262 & 0.304 & 0.295 & 0.308 & 0.100 & 0.168 & 0.140 & 0.136\\
\hline
\multicolumn{2}{c|}{\textbf{Average}}              & \red{\textbf{0.700}} & 0.677               & 0.352   & 0.159   & 0.289 & 0.328 & 0.327 & 0.340 & 0.133 & 0.181 & 0.139 & 0.139\\
\bottomrule
\multicolumn{2}{c|}{\textbf{Runtime(s) RN50}} & 126.1 & 45.5 & 129.6 & 133.5 & 38.9 & 69.2 & 69.5 & 39.3 & 454.3 & 38.3 & 38.5 & 38.5 \\
\multicolumn{2}{c|}{Std}                      & 38.2  & 5.9  & 34.2  & 39.6  & 2.9  & 5.4  & 5.5  & 2.9  & 31.0  & 2.7  & 2.9  & 2.8 \\
\hline
\multicolumn{2}{c|}{\textbf{Runtime(s) Alex}} & 104.5 & 23.0 & 118.6 & 107.1 & 16.2 & 48.8 & 47.9 & 16.8 & 435.4 & 16.8 & 17.5 & 17.6 \\
\multicolumn{2}{c|}{Std}                      & 30.0  & 4.5  & 67.7  & 31.1  & 1.5  & 4.8  & 4.8  & 1.6  & 36.1  & 1.6  & 2.2  & 2.2  \\
\bottomrule
\end{tabular}

\end{table*}

\begin{figure*}[htp]
  \centering
  \includegraphics[width=0.95\textwidth]{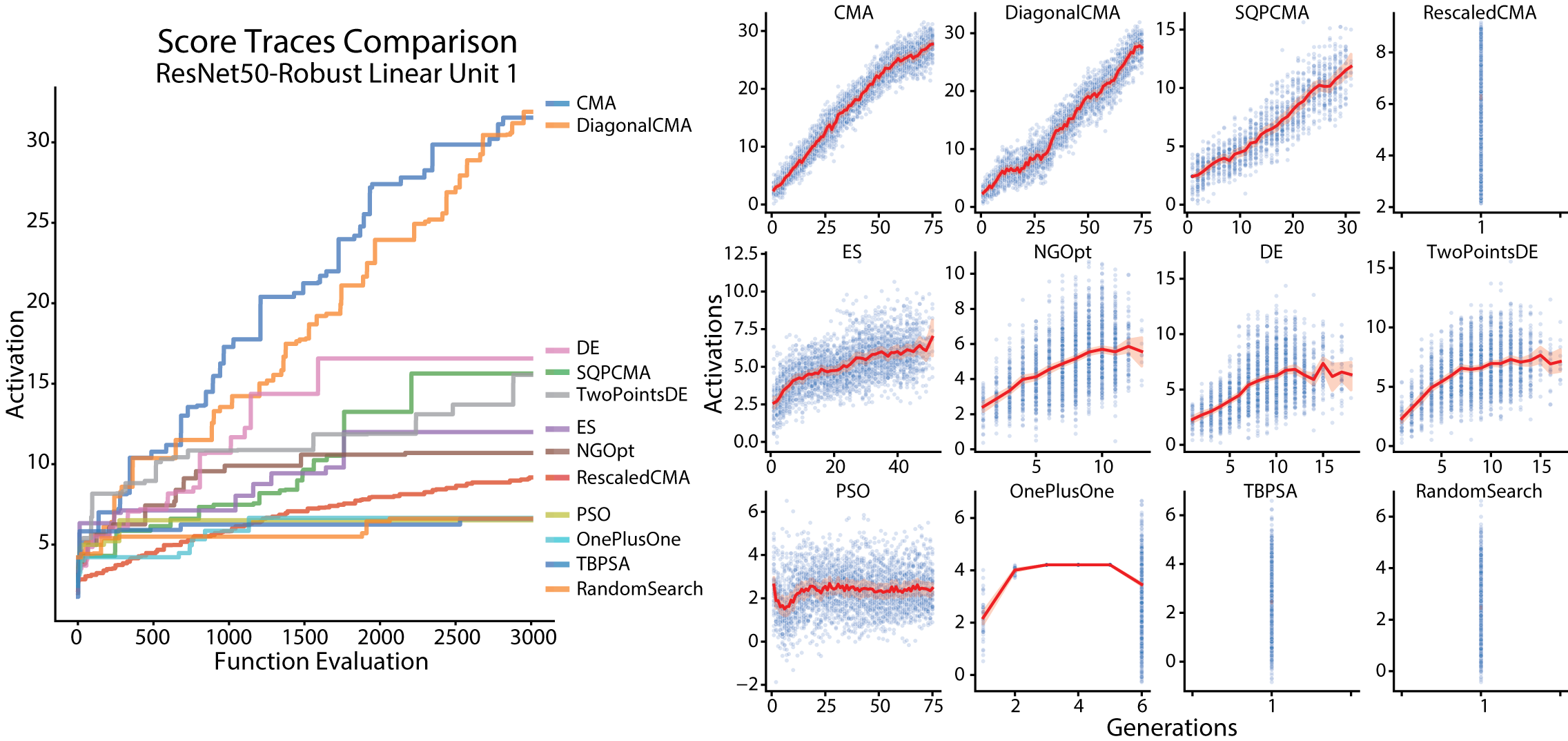}
  \caption{Score Traces Comparison of Nevergrad Optimizers with an Example Unit. \textmd{\textbf{Left}: Cumulative max score as as a function of number evaluation for each optimizer. \textbf{Right}: Scores of images plotted as a function of generations for each optimizer. The red solid trace and shading represented the mean and sem of scores each generation. Note the vast difference in the total generation number among optimizers. As the highest performing optimizers, CMA and DiagonalCMA also used the largest number of generations. All runs were performed with the unit 1 (0-indexing) in the final Linear layer of ResNet50-robust, without noise.}}
  \Description{}\label{fig:ng_optim_trajectory}
\end{figure*}

\begin{figure*}[htp]
  \centering
  \includegraphics[width=0.95\textwidth]{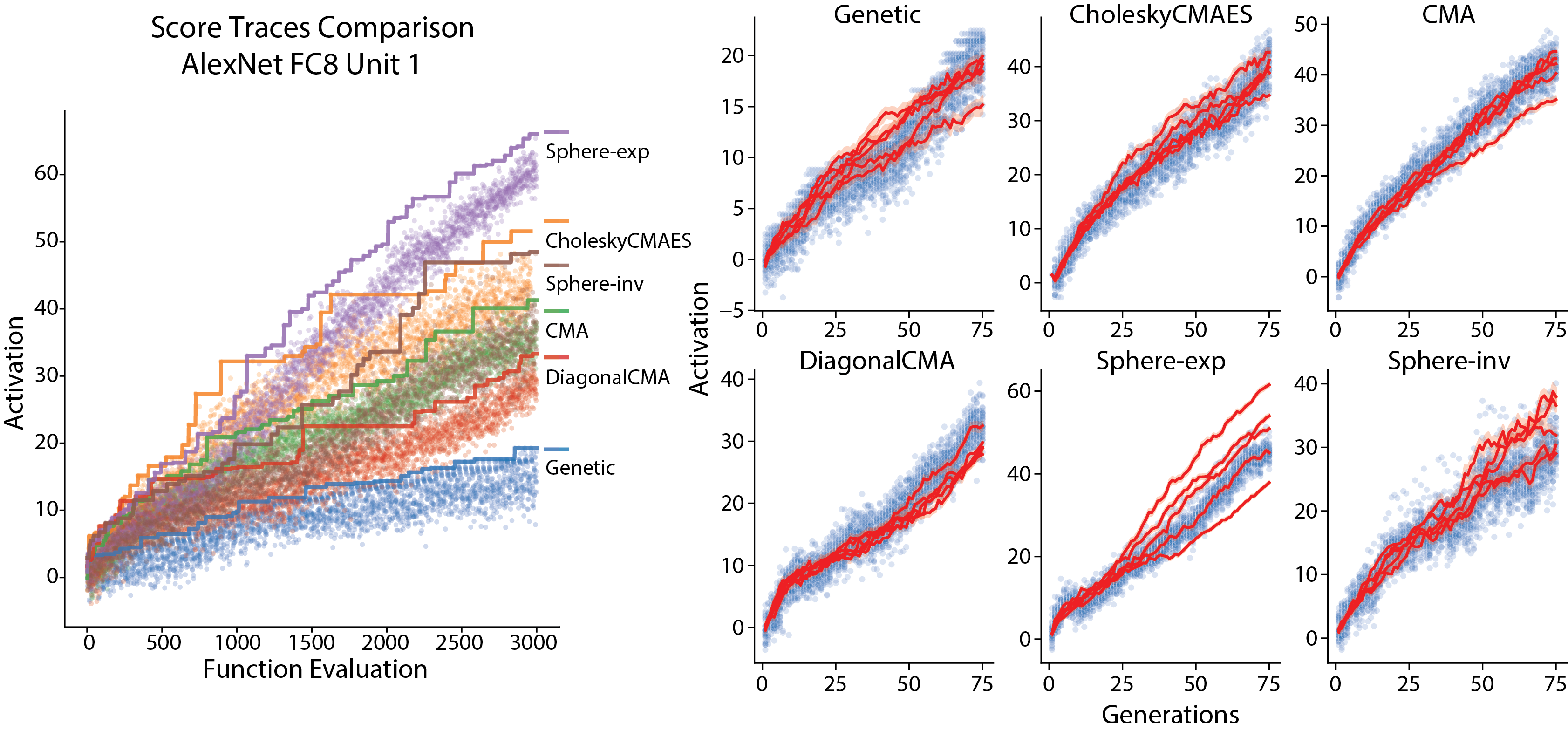}
  \caption{Score Traces Comparison of GA, CMA and Spherical Optimizers with an Example Unit. \textmd{\textbf{Left}: Cumulative max score as as a function of number of evaluation (image presentation), with scores of individual images  showed as the cloud with same color. \textbf{Right}: Scores as a function of generations. The red solid trace and shading represented the mean and sem of scores each generation, 5 runs with different random seeds were showed. All runs were performed with the unit 1 (0-indexing) in the final fc8 layer of AlexNet, without noise.}}
  \Description{}\label{fig:cmagasphere_optim_trajectory}
\end{figure*}

\begin{figure*}[htp]
  \centering
  \includegraphics[width=0.6\textwidth]{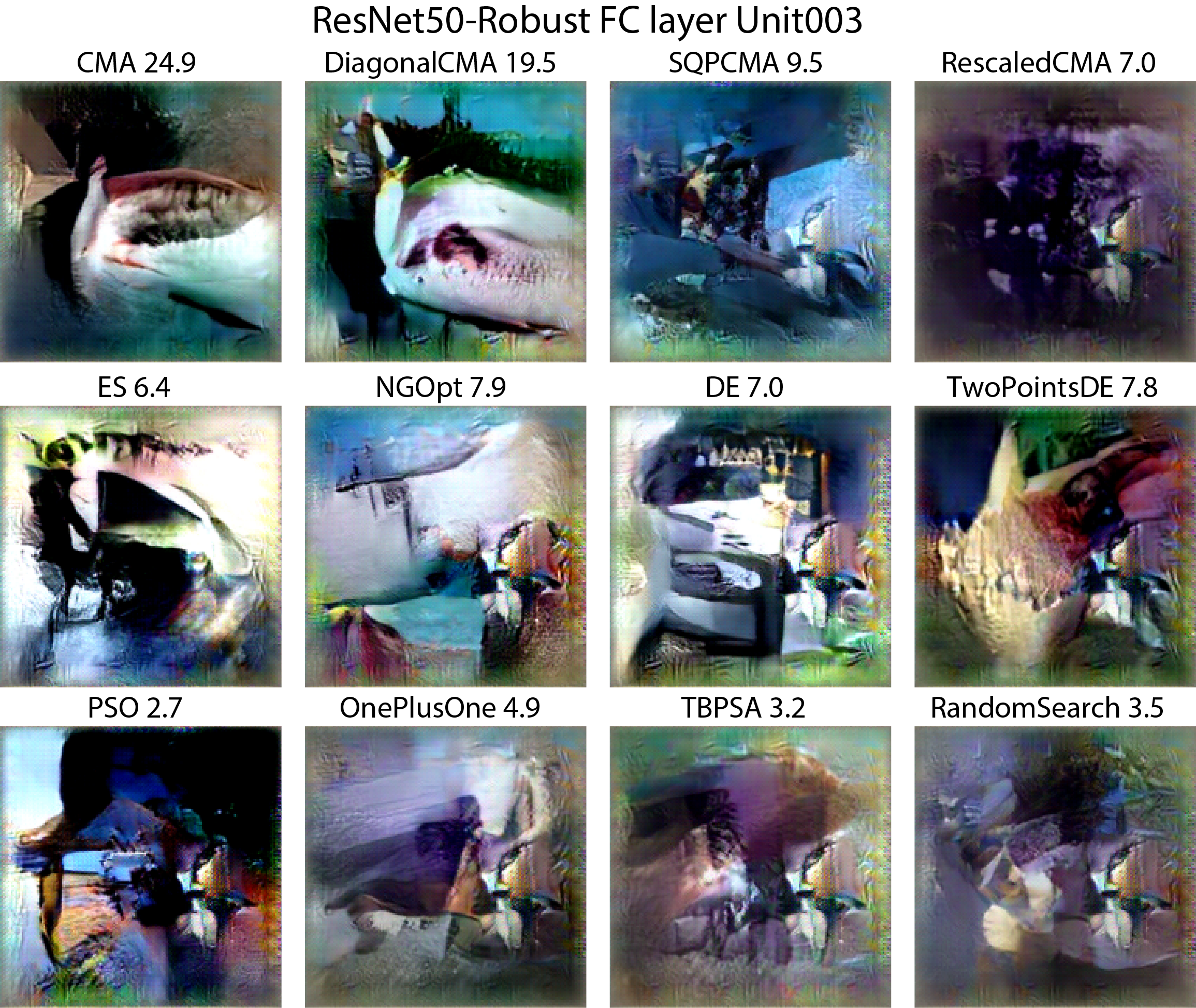}
  \caption{Comparison of Generated Images from the Code with Highest Score for each Optimizers \textmd{The objective function was the noise-free unit 3 in fc layer of ResNet50-Robust which corresponds to ImageNet label tiger shark. Arguably, the images with highest scores (noted in title) found by CMA and DiagonalCMA contained features we associated with "tiger shark", which was consistent with the function of the target unit -- to classify images as tiger shark.}}
  \Description{}\label{fig:nevergrad_proto_cmp}
\end{figure*}

\begin{figure*}[htp]
  \centering
  \includegraphics[width=.99\textwidth]{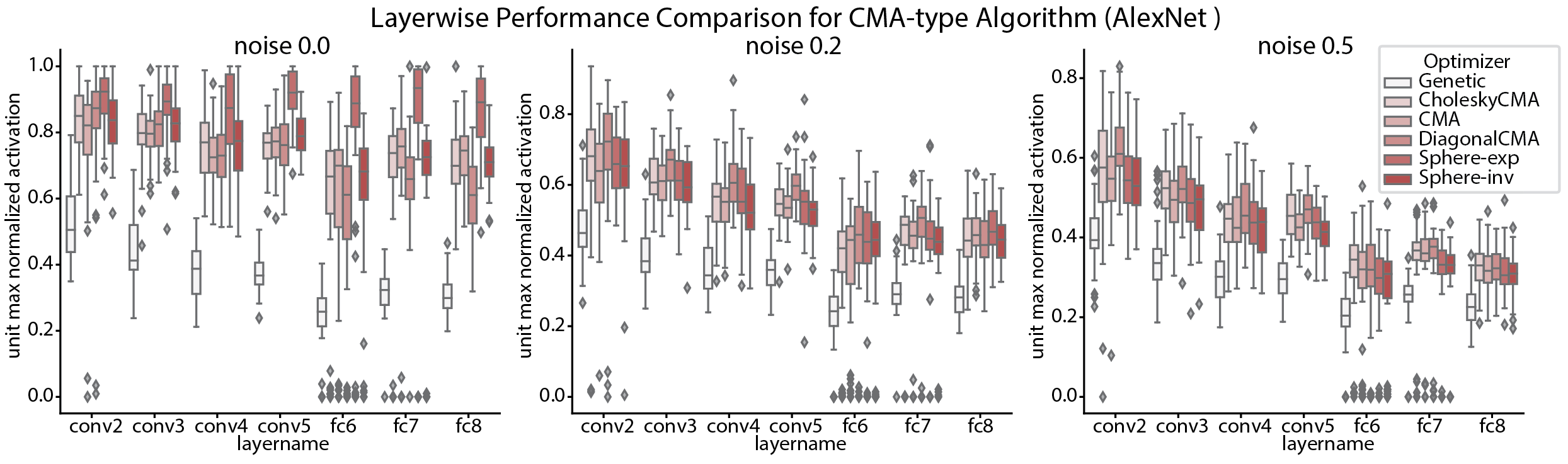}
  \caption{Layer-wise performance comparison of CMA-style algorithms. \textmd{Plotted with the same data as Tab. \ref{tab:CMA_perform_cmp}. Box and inset represent quartiles of the normalized scores for units in each layer.}} 
  \Description{}\label{fig:cma_layerwise}
\end{figure*}

\begin{figure*}[htp]
  \centering
  \includegraphics[width=.99\textwidth]{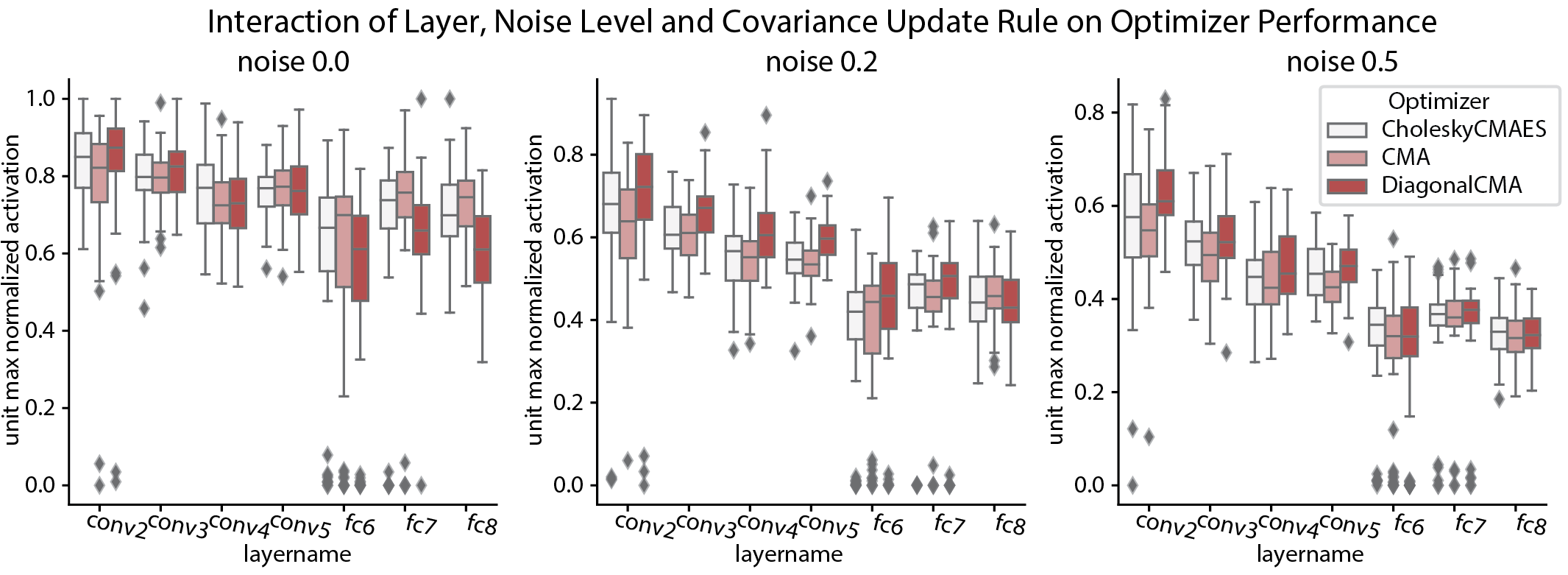}
  \caption{Interaction between unit depth, noise level and covariance update rules on optimizer performance in CNN. \textmd{Plotted with the same experimental data as Tab. \ref{tab:CMA_perform_cmp}. Same plotting convention as Fig. \ref{fig:cma_layerwise}. The statistical test for interactions showed in Tab. \ref{tab:perf_lm_anova_test}.} }
  \Description{}\label{fig:cma_partial_layerwise}
\end{figure*}

\subsection{Detailed Methods for \textit{in vivo} Evolution Experiments}\label{sec:invivo_exp_method}
Here we detailed the method by which we conducted \textit{in vivo} Evolution experiments, same as that in \citep{Rose2021Ponce}. 

Two macaque monkeys (A,B) were used as subjects with Floating Multielectrode Array (FMA) implanted in their visual cortices, in V1/V2, V4 and posterior IT. In an electro-physiology experiments, the electric signal recorded in each electrode (i.e. channel) will be processed and the spikes were detected from it using an online spike sorting algorithm from Plexon. These spikes represented the output from a few neurons or a local population in the visual cortex.

In an \textit{in vivo} session, we first performed a receptive field mapping experiment. An image was rapidly (100ms duration) showed in a grid of positions in the visual field. The spike times following the stimuli onset will be binned into a histogram, i.e. post-stimulus time histogram (PSTH). We measured the spatial extant where the image evoked neuronal responses above the baseline level. 
Based on the PSTH of the recorded units, we selected a responsive unit, with a well-formed receptive field as our target unit. This was the \textit{in vivo} counterpart of the unit we selected in CNN \textit{in silico}.

Finally, we performed the Evolution experiment. During the experiment, the images would be presented to the animal subjects on the screen in front of them, centered at the receptive field found previously; each image were presented 100ms followed by a 150ms blank screen. The spike count in $[50,200]$ ms time window after the stimulus onset was used as the score for each image $r_i$. A set of 40 texture images from \citep{freeman2011metamers} were inverted and then generated from the generator $G$; they were used as the initial generation $\{z_0^{(i)}\}$. After the neuronal responses to all images in a generation were recorded, the latent codes $\{z_i^{t}\}$ and recorded responses $\{r_i\}$ were sent to the optimizer, which proposed the next set of latent codes $\{z_i^{t+1}\}$. These codes were mapped to new image samples $\{G(z_i^{t+1})\}$ which were showed to animal subjects again. This loop continued for 20-80 rounds until the activation saturated or the activation didn't increase from the start. When we compared two optimizers (CMA vs GA in Sec. \ref{sec:cma_ga_invivo}), we ran them in parallel, i.e. interleaving the images that were proposed by the two optimizers. This ensured that the performance difference of the two optimizers were not due to change in signal quality or in electrode position.

\begin{figure}[!htp]
  \centering
  \includegraphics[width=0.35\textwidth]{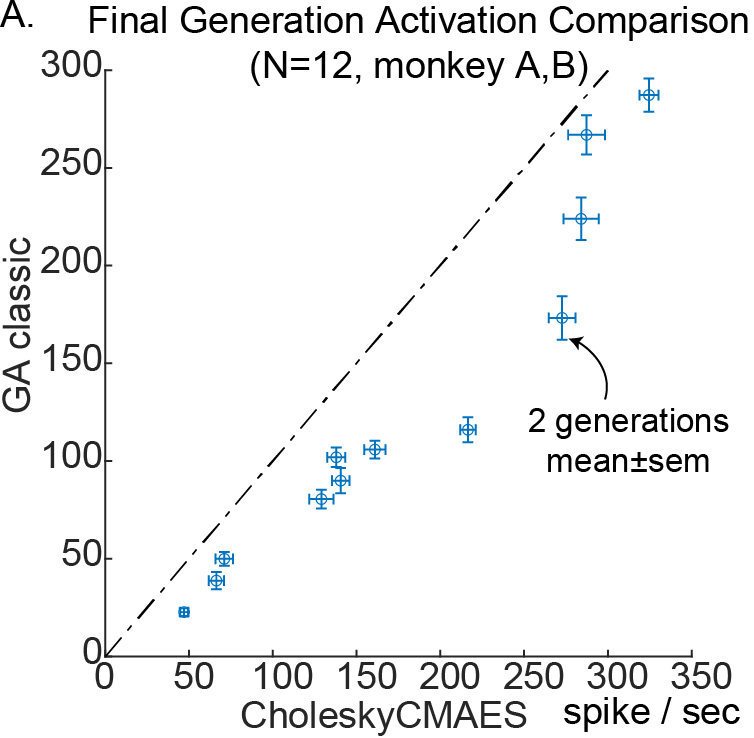}
  \caption{Raw Activation Achieved by CMA and GA in \textit{in vivo} Comparison \textmd{\textbf{A.} Comparing the response amplitude (firing rate in 50-200ms with baseline rate subtracted) in the last 2 generations for the two optimizers, mean $\pm$ sem were showed. }}\label{fig:GA_CMA_cmp_supp}
  \Description{}
\end{figure}

\begin{figure}[!htp]
    \centering
    \includegraphics{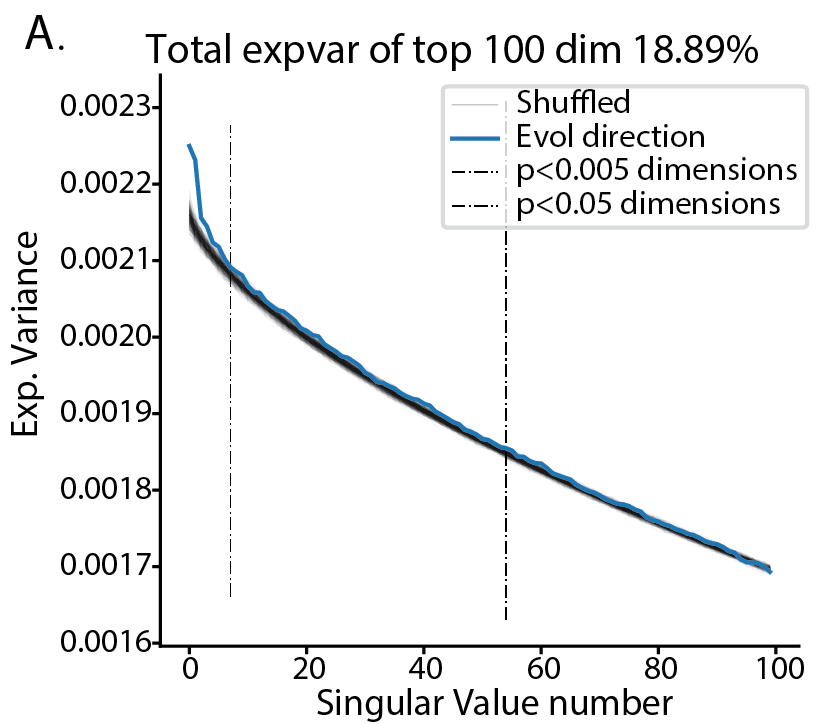}
    \caption{Collection of Evolution Trajectories were Lower Dimensional. \textmd{\textbf{A.} Comparing the SVD of the collection of \textit{in silico} evolution directions and the collection of shuffled directions.} }\label{fig:hessian_align_supp}
\Description{}
\end{figure}

\begin{figure}[!htp]
  \centering
\vspace{-5pt}
  \includegraphics[width=0.47\textwidth]{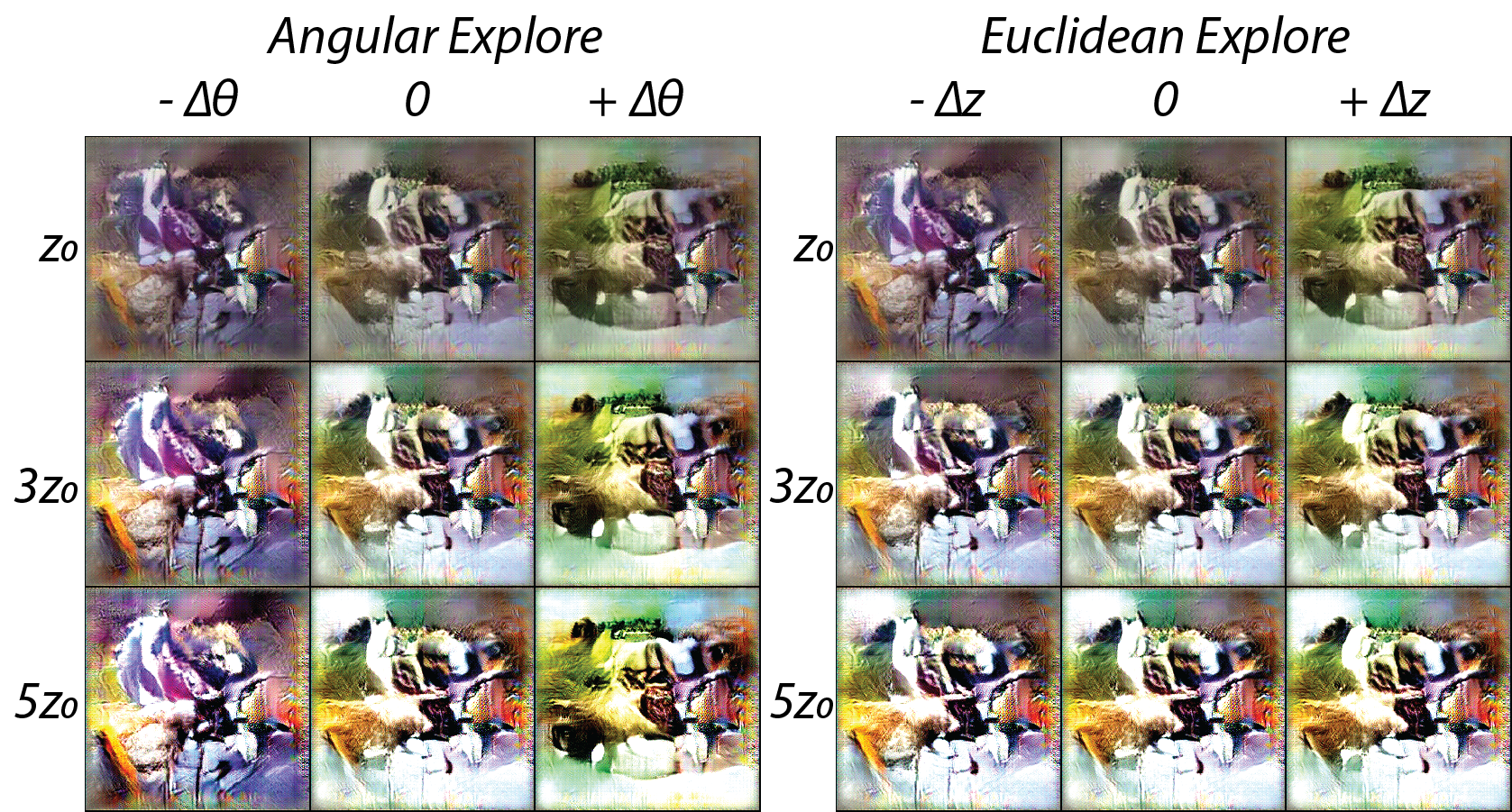}
  \caption{Example of Exploration with fixed distance versus fixed angle. \textmd{Images generated from a random vector $z_0$ and its scaled versions $3z_0,5z_0$ were showed in the center columns. Left panel visualized the perturbed vectors by the same angle $\Delta \theta$; Right panel showed the perturbed vectors by a fixed vector $\Delta z$. The first rows in two columns the same. In left panel, the columns are more similar; while in right panel, the images in the 2nd and 3rd rows were more similar. We argued that with perturbation of the the same euclidean length, the effect on image will decrease for a base vector of larger norms. But perturbation of the same angle will result in similar perceptual change regardless of the norm of the base vector. Also see illustration in Fig. \ref{fig:supp_GANgeometry}A}}
  \Description{}\label{fig:example_ang_eucl}
\vspace{-10pt}
\end{figure}

\begin{figure}[!htp]
  \centering
\vspace{0pt}
  \includegraphics[width=0.40\textwidth]{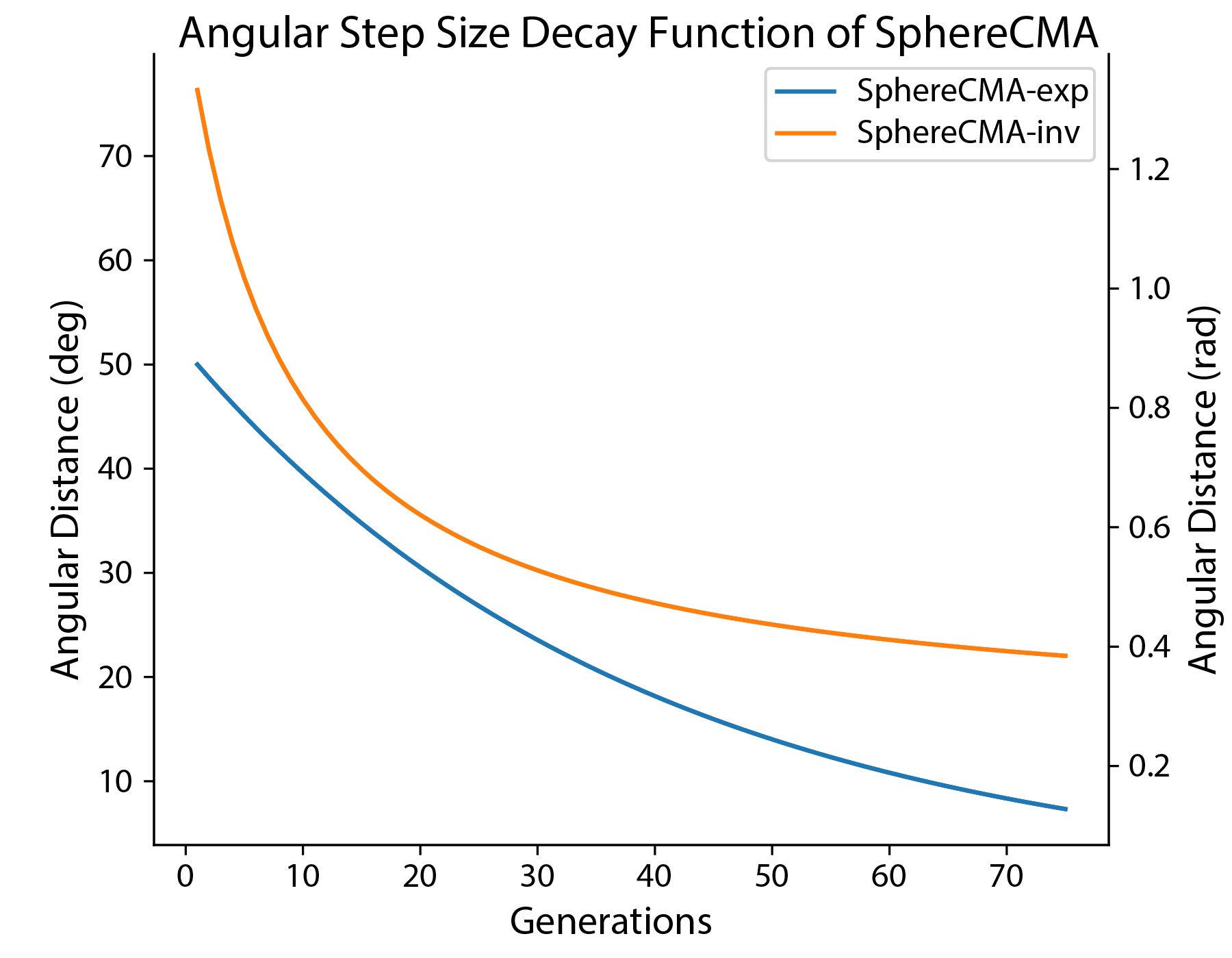}
  \caption{Angular step size decay curves for SphereCMA-Exp and SphereCMA-Inv.}
  \Description{}\label{fig:Sph_step_tune}
\vspace{-5pt}
\end{figure}

\section{Geometry of the Generative Image Manifold}\label{sec:GAN_geometry}
\begin{figure*}[htp]
  \centering
\vspace{0pt}
  \includegraphics[width=1.0\textwidth]{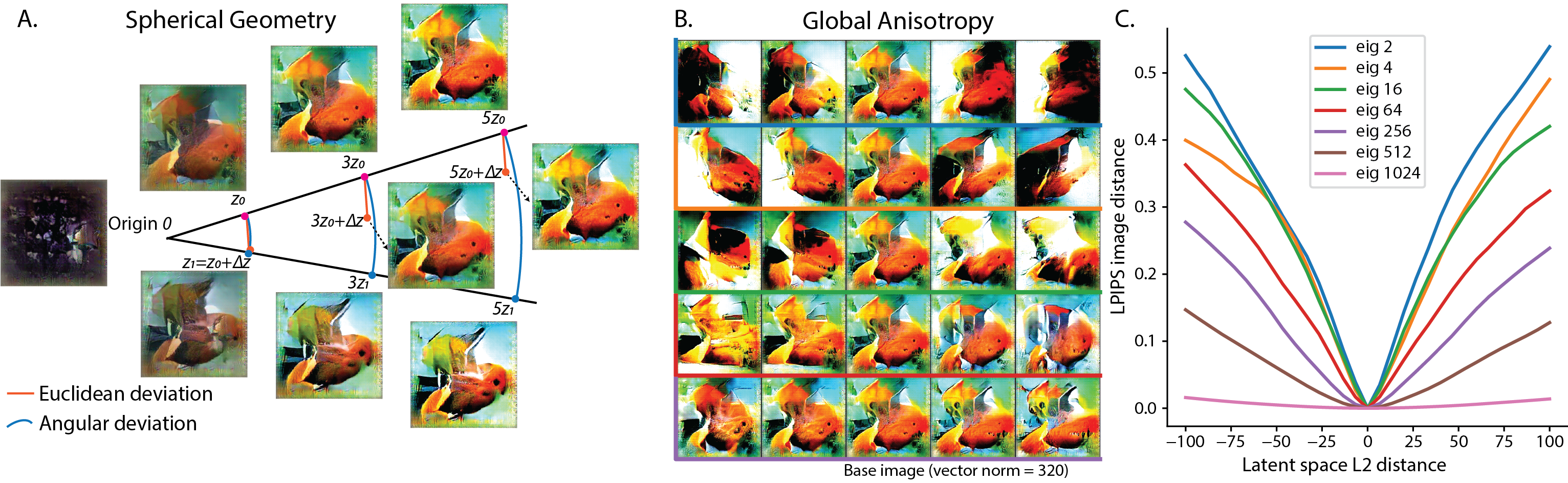}
  \caption{Spherical Geometry and Anisotropy of the Generative Image Space. \textmd{\textbf{A.} illustrates the spherical geometry of generative space. \textbf{B.} illustrates the anisotorpic structure of the generative space. Each row visualizes the effect on generated image by perturbing the reference code $z_0$ along an eigenvector by L2 distances $d=-100,-50,0,50,100$, i.e. $z_0+du_k$. The rows correspond to five eigenvectors (number $k=2,4,16,64,256$), showed in descending order of their eigenvalues. \textbf{C.} The LPIPS image distance as a function of the Euclidean distance along each eigen dimensions. The images in \textbf{B} visualized the perceptual meaning of the LPIPS distances in \textbf{C}. Structure of \textbf{B.C.} was inspired by Fig.1 in \citep{Wang2021Geometry}.}}
  
  \Description{}\label{fig:supp_GANgeometry}
\vspace{-5pt}
\end{figure*}

As showed in \citep{Wang2021Geometry}, the deep neural network image generators usually have a highly anisotropic latent space. We visualized it in Fig. \ref{fig:supp_GANgeometry} B,C. When travelling the same distance along an eigenvector, it induced much larger perceptual change in the image when the eigenvalue was larger; and eigenvectors like the rank 1000 one (Fig. \ref{fig:supp_GANgeometry}B) merely changed the image in the range tested. Thus these lower eigenvectors formed an effective "null space" where the input was not meaningful to the generated images or the activation of the units or neuron. In later development of optimization algorithms \citep{Wang2021Geometry}, we projected out the lower eigen-dimensions and only optimize the top eigen-dimensions of the generator. This could reduce the search dimensionality to around 500d. 

As for the spherical geometry of the generative image mapping $G$, it can be seen from Fig. \ref{fig:supp_GANgeometry}, scaling the latent vectors will majorly change the contrast instead of content of the generated image. This made angular distance relevance to our generator. However, unlike anisotropy, this linearity of $G$ is not a general property of generators. For other image generators such as BigGAN, StyleGAN \citep{brock2018BigGAN,karras2019styleGAN,karras2020StyleGAN2}, scaling the latent vector will not result in a scaling in pixel space: the image content will also change. We suggest that for these generator spaces, the code norm should not be fixed but should be varied and explored during optimization. So when developing optimizers in those latent spaces, as code norm increases automatically for CMA type algorithm due to diffusion, unconstrained evolution will bias the generated image towards those images with larger code norms. 

\section{Geometry of the Tuning Function Landscape}\label{sec:funct_geomtry}
The geometry of the latent space is important for the effectiveness of optimization algorithms. More important is the geometry of the objective function in this space. As a foundation, we characterized the geometry esp. curvature of the tuning functions of \textit{in silico} units. 
Recall the tuning function of unit in the image space $f:\mathcal I\to \mathbb R$, and the image generative map $G:\mathbb R^d\to\mathcal I$. The effective function to be optimized in our problem was their composition $f\circ G:\mathbb R^d\to \mathbb R$, we measured its curvature by computing the Hessian spectra of $f\circ G$. Intuitively, this eigenvalue represented the sharpness of tuning along the eigenvector in the latent space: a larger eigenvalue corresponds to a sharper tuning and a smaller eigenvalue corresponds to a broader tuning. 

We selected 5 units from each of the 8 layers of AlexNet, and then searched for a local maximum $z^*$ for each unit using CholeskyCMAES optimizer. Next, we calculated the eigendecomposition of the Hessian $H=\frac{\partial f\circ G}{\partial z\partial z}|_{z^*}$ evaluated at the local optima. We borrowed the technique from \citep{Wang2021Geometry}: we represented the Hessian matrix implicitly by constructing a Hessian vector product (HVP) operator, and then applied Lanczos algorithm on this linear operator to extract the top 2000 eigen-pairs with the largest absolute eigenvalues. We examined the absolute eigen spectra, normalized by the largest absolute eigenvalue. This was reasonable since the overall scale of eigenvalues was related to the scale of activation of the unit, which differed from layer to layer. 

We found that all the spectra were highly "ill-conditioned" spanning eigenvalues of 4 to 9 orders of magnitude. This was consistent with the findings in \citep{Wang2021Geometry}. The image generative network $G$ is already a highly ill-conditioned mapping, such that some latent dimensions affects the image much more than other dimensions. Since the tuning function $f\circ G$ in the latent space is a composite of $f$ and $G$, the Jacobians of the two mappings multiply, so it inherited this ill-condition of the generative map. 

\begin{figure*}[!htp]
  \centering
  \includegraphics[width=0.48\textwidth]{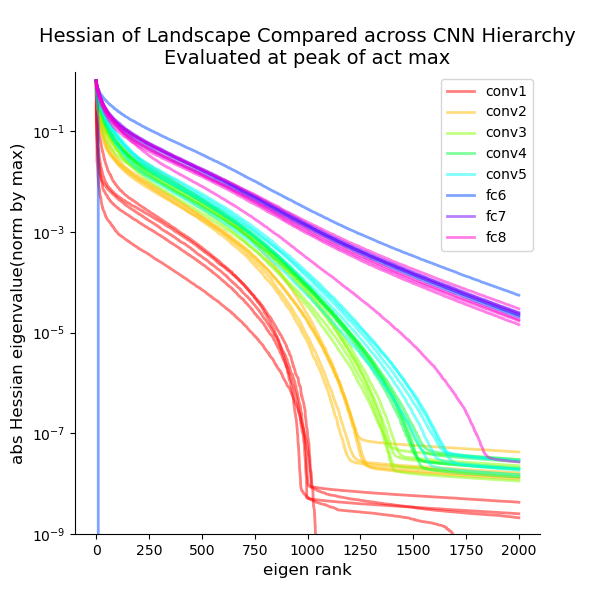}
  \includegraphics[width=0.48\textwidth]{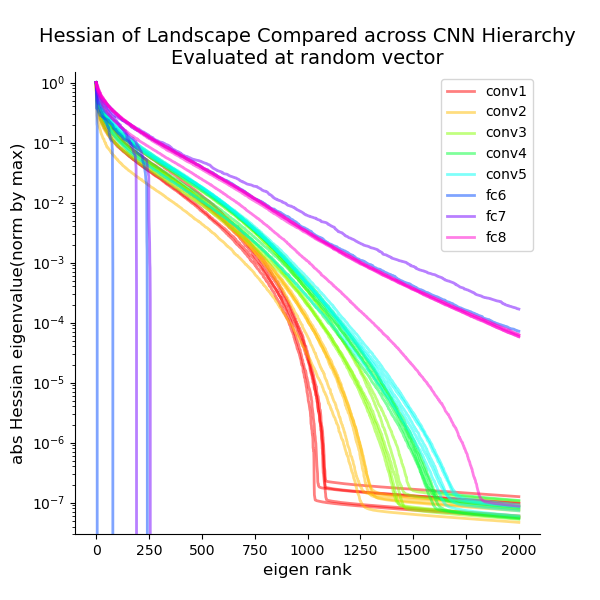}
  \caption{Hessian Spectrum of Units Across AlexNet Hierarchy. \textmd{\textbf{Left.} Evaluated at the peak i.e. the final generation latent code of a CMAES search. \textbf{Right.} Evaluated at random initialized vectors. It could be appreciated that conv1 had a sharp decay in its spectra while the spectra of conv2, conv3, conv4 units got gradually more high dimensional. Fully connected layers fc6-fc8 all exhibited higher dimensional and slower decay spectra.}}\label{fig:CNN_Hess_Geometry}
  \Description{}
\end{figure*}

More interestingly, as we compared the spectra of tuning landscapes across layers, we found that the spectra decayed rapidly for units in conv1 and decayed slower going up the hierarchy from conv2 to conv5, and became much slower for units in fully connected layers fc6-fc8 (Fig. \ref{fig:CNN_Hess_Geometry}). We found similar spectral result when we evaluated the Hessian at random vectors $z_{rnd}$ instead of the peaks found by optimizers $z^*$, confirming that this was a general phenomenon of the landscape. Though we noted that, the spectra of fc layer units precipitated sometimes, presumably because the unit was not active enough. 

We interpreted this as units in early convolutional layers were tuned for very few dimensions in the latent space, and they were relatively invariant for a large subspace. In contrast, the units deeper in the hierarchy were tuned for a larger dimensional space, and were selective for more specific combination of features. 

This observation was relevant to the performance of optimizers with different covariance update rules (Tab. \ref{tab:CMA_perform_cmp}). For units in earlier conv layers (e.g. conv1-5), there was a large invariant subspace, so the peaks of optimization were easy to find, so optimizers don't need to find or model the exact tuning axis. In contrast, for units that tuned for hundreds-thousands of dimensions (fc6-8) all these axes need to be be jointly optimized to achieve the highest activation. Because of this, we reasoned that the diagonal scaling of covariance matrix may not work well in the deeper layers like fc6-8. This could be one explanation for why DiagonalCMA perform better than CMA in earlier layers, but not as well in deeper layers (Tab. \ref{tab:perf_lm_anova_test}).

\section{Additional Analysis}
\subsection{Dimensionality of the Collection of Evolution Trajectories}
We tested if the collection of evolution trajectories exhibited a lower dimensionality than shuffled controls. We collected the evolution trajectories of $N=1050$ runs, across all noise level and layers and their PC structure. Since we observed that the mean trajectories could be well represented by their top PCs, we represented each search trajectory by a single vector, evolution direction $\zeta$. We computed this direction $\zeta$ by projecting the mean latent code of final generation $\bar{z_T}$ onto the top 5 PCs space $\zeta=\mathcal P_{PC1:5}\bar{z_T}$ (which explained over 88\% of the variance of the mean trajectory). 
We shuffled the entries of these vectors to form a control collection of trajectories $z_{shuffle}$, preserving their vector norms. We shuffled 500 times to establish statistic significance. 

Generally the collection of evolution directions were \textit{not low dimensional}, with top $100$ dimensions accounting for only $18.9$\% of variance (Fig. \ref{fig:hessian_align_supp} A). But comparing to the shuffled counterparts, the evolution trajectories resided in a (slightly) lower dimensional space: The first 7 singular values were larger than the shuffled counterpart ($p<0.005$, Fig. \ref{fig:hessian_align_supp} A). This signature suggested that the search trajectories exhibited some common lower dimensional structure.


\begin{figure*}[!htp]
  \centering
  \includegraphics[width=\textwidth]{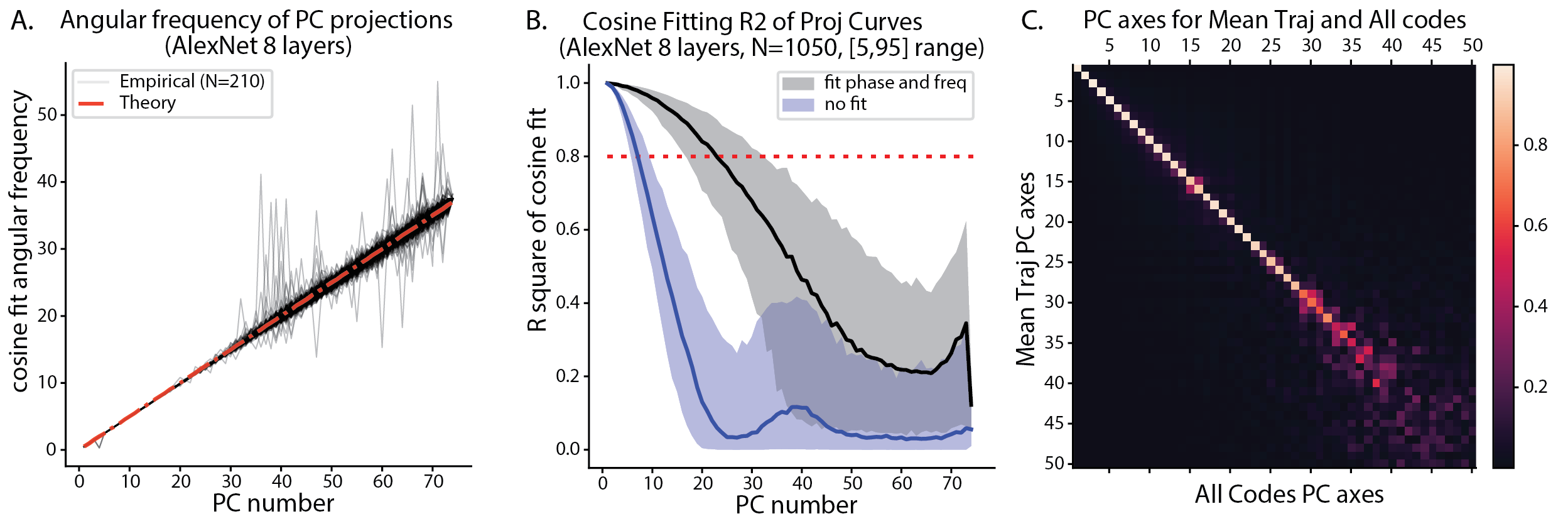}
  \caption{Sinusoidal Geometry of Evolution Trajectory as Signature of Random Walk (continued). \textmd{\textbf{A.} Angular frequency $\omega$ of the cosine fitting as a function of PC number $k$, with the theory value $\omega=k/2$. $N=210$ evolutions were plotted. \textbf{B.} $R^2$ of cosine function for the $k$th PC projection. Central line showed the median, shaded area showed the $[5,95]$\% confidence interval. \textbf{C.} Top PCs for the mean trajectory and top PCs for all latent codes were well aligned. }}\label{fig:sinusoidal_supp}
  \Description{}
\end{figure*}

\begin{figure*}[!htp]
  \centering
  \includegraphics[width=0.95\textwidth]{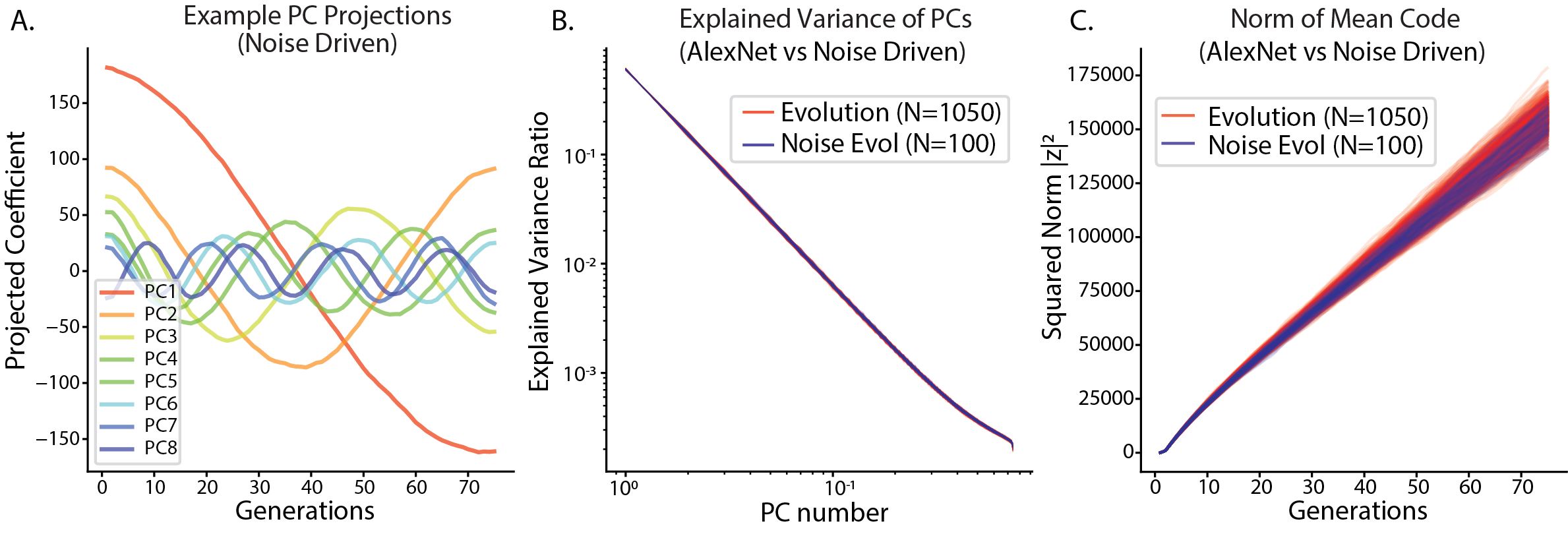}
 
  \caption{PCA of Noise-Driven Evolution also Exhibits Sinusoidal Structure \textmd{\textbf{A.} Mean search trajectory projected onto the first 8 PCs of a noise-driven Evolution. \textbf{B.} Explained variance as a function of PC number $k$ for noise driven evolution and real evolution coincided well. The existence of driving force seems not to have a large effect on the spectrum of Evolution trajectories. \textbf{C.} Code norm during evolution for noise driven and AlexNet units driven evolution. For both, the squared norm scaled linearly with the Generation. At the final generation, the norm of latent codes was higher for AlexNet driven evolution: Mean norm in evolution experiments was 394.80$\pm$7.05 (mean $\pm$ std) in contrast to the noise-driven control 390.96$\pm$7.09 ($t=5.257,p=1.7\times10^{-7},df=1151$). The difference in norm is significant, but the effect size is small. 
}}\label{fig:sinusoidal_noise_supp}
  \Description{}
\end{figure*}

\subsection{PCA of Evolution Trajectory Formed by All Codes also Exhibited Sinusoidal Structure}
In addition to the analysis we did to the mean search trajectory as in Sec. \ref{sec:sinusoidal}, we considered how individual latent codes sampled during Evolution fit into the global geometry of the trajectory. 
We stacked all the latent codes $\{z^{(i)}_{t}\},t=1...,T,i=1,...n_P$ sampled by the CholeskyCMA during Evolution into a $T\cdot n_P$-by-$d$ matrix $\tilde X=[z^{(1)}_1,z^{(2)}_1,...,z^{(n_P)}_T]$ ($T\cdot n_P=3000$) and applied PCA to decompose it. We found that the main structure of the search trajectory remained intact: the codes projected onto top principal axes were still sinusoidal curves but with dispersion. 
We compared the PC vectors computed from all latent codes, and those from the mean trajectory by calculating their the inner product or cosine angles. We found that the top 30 PC vectors of two basis were matched to a high degree: 28 PC vectors of the mean trajectory had cosine angle > 0.8 with the PC vector of the same rank for all latent codes (Fig. \ref{fig:sinusoidal_supp} C). This showed that the global structure of all sampled codes coincided with the mean search trajectory, while the individual codes contributed to local variability around the mean search trajectory.

\subsection{2D PC Projections of Evolution Trajectory were Lissajous Curves}
\begin{figure*}[!htp]
  \centering
  \includegraphics[width=0.9\textwidth]{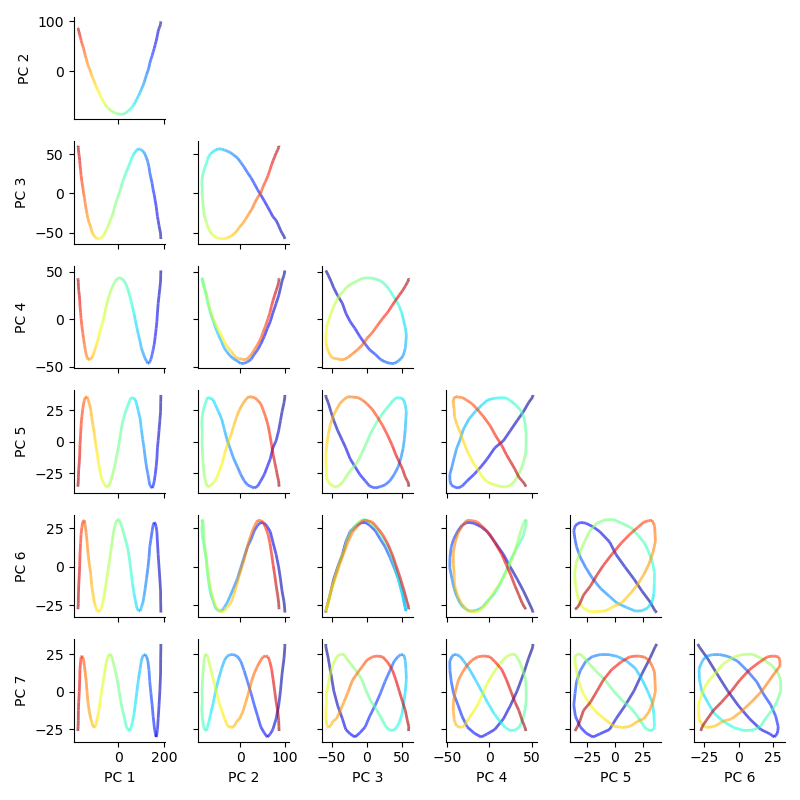}
  \caption{Lissajous Curves formed by PC Projection of Evolution Trajectory \textmd{Blue to red color gradient represents the progression of time, from first to the last step. This is obtained from an example Evolution trajectory with AlexNet FC8 unit 0 as objective and CholeskyCMA as optimizer.  (compare Fig. 15 in \citep{antognini2018pcaHighDimRandomWalk}).}}\label{fig:LissajousGrid}
  \Description{}
\end{figure*}

\begin{figure*}[!htp]
  \centering
  \includegraphics[width=0.9\textwidth]{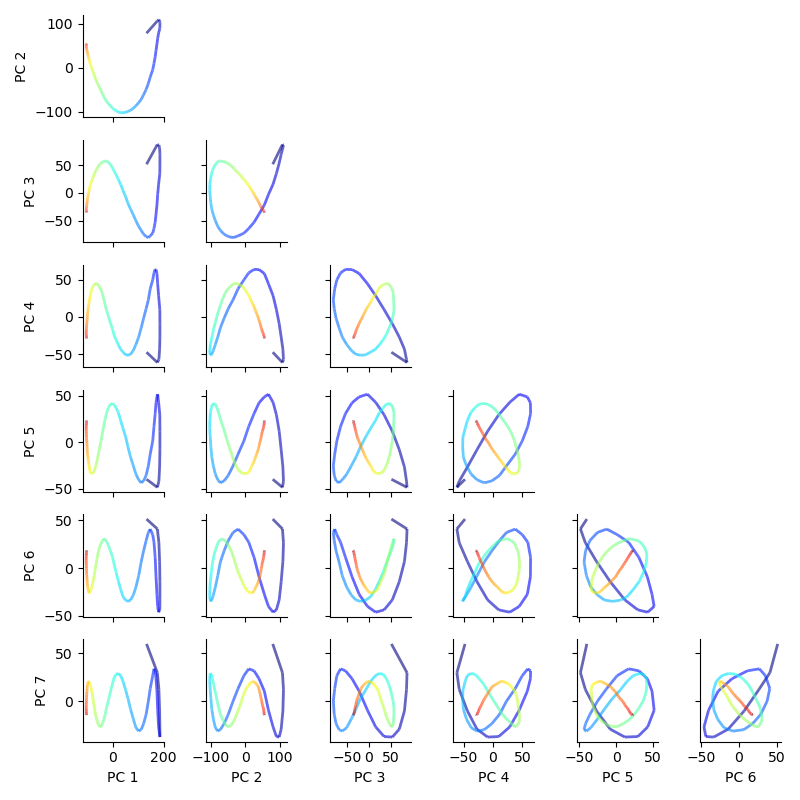}
  \caption{Lissajous Curves formed by PC Projection of Evolution Trajectory Driven by SphereCMA \textmd{Blue to red color gradient represents the progression of time, from first to the last step. This is obtained from an example Evolution trajectory with AlexNet FC8 unit 0 as objective and SphereCMA as optimizer. (compare Fig. \ref{fig:LissajousGrid}).}}\label{fig:LissajousGrid_sphereCMA}
  \Description{}
\end{figure*}

In our work, we consistently observed that the projection of original CMA, CholeskyCMA and Diagonal CMA evolution trajectory onto top PC axis was sinusoidal curve. One interesting way to visualize these trajectories is to project it onto 2d PC subspaces. As expected, we saw the Lissajous curve formed by the projection (Fig. \ref{fig:LissajousGrid}). We noted that the Lissajous curves decreased its amplitude towards the end (red), so the red end didn't overlap perfectly with the blue end. This is in line with a decreasing step size during evolution. As a comparison, we also performed PCA and 2d projection with optimization trajectories driven by SphereCMA algorithm (Fig. \ref{fig:LissajousGrid_sphereCMA}). We can see, it deviated more from the sinusoidal structure predicted by random walk, indicating a different kind of search process on the sphere. 

\section{Definition of Subroutines used in \texttt{SphereCMA}}\label{sec:SphereCMA_subrountines}
In this section we define the subroutines we used in the \texttt{SphereCMA} algorithm: \texttt{ExpMap}, \texttt{RankWeight}, \texttt{SphereExtrapolate}. 

\begin{algorithm}[H]
\caption{ExpMap}
\label{alg:expmap}
\begin{algorithmic} 
\Require Base vector $m$, tangent vector $v$, angle $\mu$ to travel from $m$ towards $v$.
\Ensure $\|m\|>0$, $m^Tv=0$
\State $R \gets \|m\|$
\State $\hat m\gets m/R$
\State $\hat v\gets v/\|v\|$
\State $u\gets R\cdot(\hat m\cos \mu + \hat v\sin \mu$) 
\State \Return $u$
\end{algorithmic}
\end{algorithm}

\begin{algorithm}[H]
\caption{SphereExtrapolation.\\
This is the famous SLERP (Spherical linear interpolation) algorithm.}\label{alg:SphExtrapolation}
\begin{algorithmic} 
\Require Base vector $m$, end vector $p$, ratio $t$ controlling interpolation or extrapolation.
\Ensure $\|m\|=\|p\|$
\State $\hat m\gets m/\|m\|$
\State $\hat p\gets p/\|p\|$
\State $\theta=\arccos (\hat m^T\hat p$)
\State $q\gets (\sin((1-t)\theta)m + \sin(t\theta)p) / \sin\theta$
\State \Return $q$
\end{algorithmic}
\end{algorithm}

\begin{algorithm}[H]
\caption{RankWeight}\label{alg:rankweight}
\begin{algorithmic} 
\Require Array of scores $r=[r_1...r_B]$, population size $B$
\Require Parameter: cutoff number $K=B/2$
\State Initialize raw weights $\tilde w$ as an array of length $B$.
\State Top $K$ candidates get positive weights. For $k=1$ to $K$, $\tilde w_k\gets log(K+1/2)-log(k)$
\State Other candidates get no weights. For $k=K+1$ to $B$, $\tilde w_k\gets 0$
\State Normalize all weights to sum to one. $\tilde w_i\gets \tilde w_i/\sum_i\tilde w_i$
\State Find rank $N_i$ of each scores $r_i$ (e.g. largest score has rank 1.)
\State Fetch the raw weight by the rank of each score. For $i=1$ to $B$, $w_i\gets \tilde w[N_i]$ 
\State \Return $w$
\end{algorithmic}
\end{algorithm}

\end{document}